\def\vtcpreprintversion{1}
\def\vtcpreprintauthors{%
      \makebox[\dimexpr\textwidth-2\tabcolsep\relax][c]{%
            \begin{tabular}{c@{\hspace{3em}}c}
                  \rule{0pt}{3ex}\textbf{Florian Le Bronnec}        & \textbf{Rio Yokota} \\
                  {\normalfont\ttfamily florian.lebronnec@riken.jp} &
                  {\normalfont\ttfamily rio.yokota@riken.jp}                              \\[1ex]
                  \multicolumn{2}{c}{\normalfont RIKEN Center for Computational Science, Tokyo, Japan}
            \end{tabular}
      }%
}
\documentclass{article}

\newif\ifvtcpreprint
\ifdefined\vtcpreprintversion
      \vtcpreprinttrue
\else
      \vtcpreprintfalse
\fi

\usepackage{iclr2027_conference,times}
\usepackage{graphicx}
\usepackage{subcaption}
\usepackage{amsmath,amssymb}
\usepackage{booktabs}
\usepackage{float}
\usepackage{enumitem}
\usepackage{rotating}
\usepackage{tabularx}
\usepackage{hyperref}
\usepackage{url}
\definecolor{resultpositive}{HTML}{006D5B}
\definecolor{resultsmall}{HTML}{9A6700}
\definecolor{resultneutral}{HTML}{5F6368}
\newcommand{\tableheat}[3]{%
\begingroup
\setlength{\fboxsep}{1pt}%
\colorbox{#1!#2}{\makebox[3em][r]{\strut #3}}%
\endgroup}

\title{Evaluating Multiple LLM Generations with Validated Task Coverage}

\ifvtcpreprint
      \author{\vtcpreprintauthors}
      \iclrfinalcopy
\else
      \author{Anonymous authors}
\fi

\begin{document}

\maketitle
\ifvtcpreprint
      \lhead{Preprint}
      \vspace{-0.2in}
\fi

\begin{abstract}
      Many LLM applications are most useful when they provide several candidate outputs for comparison, validation, or combination. Predominant evaluation settings,
      however, still focus on individual outputs or reduce multiple samples to a single success or selected
      answer. This can miss whether the outputs include several genuinely different useful results.
      We introduce VTC-Bench, a five-domain benchmark for this setting, together with Validated Task Coverage (VTC) as its core evaluation quantity. The benchmark is built from carefully selected real-data tasks
      where both output quality and task-relevant distinctness can be checked automatically and reproducibly, without model-based judges. VTC measures how many distinct useful results are obtained within $k$ attempts.
      Across multiple models and inference settings, the benchmark leads to different conclusions
      from conventional evaluation: configurations that look strongest from single-draw quality are not
      necessarily those with the best coverage, and simple measures of output variation do not reliably
      recover task-relevant coverage. These results show that finite candidate sets can be evaluated directly as objects of interest, revealing differences in model behavior that are not apparent from conventional per-output evaluation. \ifvtcpreprint
            Project materials are available at
            \url{https://github.com/flbbb/validated-task-coverage}.
      \fi

\end{abstract}

\section{Introduction}

Many LLM applications benefit from generating several candidate outputs rather than committing to a
single response. A user may want alternative solutions, hypotheses, designs, or pieces of evidence
to compare, validate, or combine before making a decision
\citep{ippolito2019comparison,li2022competition,shypula2025evaluating}. In such settings, the relevant object is
not an isolated generation but the set produced after repeated attempts. The value of that set
depends not only on whether individual outputs are good, but also on whether later attempts add
new useful possibilities rather than repeat what has already been found.

Most existing evaluation, however, is still organized around the quality of individual outputs.
Benchmarks typically measure accuracy, success rate, or another per-attempt score, while multi-sample
methods such as pass@$k$, self-consistency, and best-of-$N$ ultimately reduce several generations
to a single success event, aggregate answer, or selected output
\citep{chen2021evaluating,wang2023selfconsistency,huang2025bestofn}. These objectives are appropriate
when only one final answer matters. They do not measure whether generated answers accumulate several
distinct useful outcomes.

We introduce VTC-Bench, a benchmark specifically designed for this finite-candidate-set setting, together with
Validated Task Coverage (VTC) as its core evaluation quantity. VTC asks a natural question:
after $k$ attempts, how many distinct useful outcomes have been validated? Making this quantity
measurable in realistic tasks requires more than repeated generation: each task must define what
counts as a valid output and when two valid outputs represent the same useful outcome.

The main challenge is making VTC operational in practical tasks. We therefore select settings where useful outputs can be validated automatically and mapped deterministically to task-specific useful outcomes. VTC-Bench comprises five real-data tasks built around these criteria, with reproducible scoring and no reliance on model-based judges.

With the benchmark in place, we can measure both output quality and task-relevant coverage directly
on the same set of tasks. Across multiple models and inference settings, we find
that these measurements lead to different conclusions: configurations that look strongest from
single-draw quality are not necessarily those that achieve the best coverage, and simple measures of output variation do not reliably reflect task-specific coverage. Together, our contributions are:
\begin{itemize}[leftmargin=*, itemsep=0pt, topsep=0pt]

      \item \textbf{VTC-Bench, a five-domain benchmark.}
            We construct practical tasks where both output validity and task-relevant distinctness can be checked automatically and reproducibly.
      \item \textbf{Validated Task Coverage.}
            We introduce a common framework for evaluating the distinct useful outcomes produced across repeated generations.
      \item \textbf{A comprehensive empirical study.}
            Across multiple models, inference settings, and attempt budgets, we show that
            validated coverage can lead to different model-selection conclusions from single-draw quality
            and surface-level diversity.
\end{itemize}

\section{Related Work}

\textbf{Repeated-sampling metrics.}
Pass@$k$, self-consistency, and best-of-$N$ all evaluate multiple generations, but reduce them to a
single success event, aggregate answer, or selected output
\citep{chen2021evaluating,wang2023selfconsistency,huang2025bestofn}. These methods are appropriate
when only one final answer matters. They do not measure how distinct useful outcomes accumulate
across the retained set of generations, which is the object evaluated by VTC.

\textbf{Distributional metrics.}
Prior work evaluates quality and diversity jointly by comparing generated and
reference distributions, using textual similarity measures or learned representations
\citep{pillutla2021mauve,lebronnec2024exploring,alihosseini2019jointly,guo2025linguistic}.
These approaches estimate distributional agreement from samples and generally require a
sufficiently representative reference set. Our setting asks a different question: given a fixed
task instance and a finite number of attempts, how much useful task progress has actually been
accumulated?

\textbf{Task-specific coverage.}
Recent work shares our motivation that repeated generations should be evaluated by the distinct
useful outcomes they recover.
HypoSpace \citep{chen2026hypospace} is the closest conceptual precedent: it measures recovery of an enumerable valid solution space, which requires exhaustive ground truth and therefore favors deliberately controlled tasks. \citet{shypula2025evaluating} and NoveltyBench \citep{zhang2025noveltybench} study useful diversity in open-ended generation, using program semantics and learned functional-equivalence judgments, respectively. Other studies examine useful diversity for specific applications: PluralEval \citep{mundada2026pluralism} measures coverage of distinct human viewpoints for subjective questions, while \citet{lee2025diversely} measure algorithmic diversity among correct generated programs. Both rely on model-based semantic judgments to identify these distinctions.
VTC instead formalizes finite-budget coverage: for a fixed $k$, it measures how many distinct useful outcomes are recovered across $k$ attempts. We instantiate this across heterogeneous practical tasks using automatic validation and deterministic task-specific outcome mappings.

\section{Validated Task Coverage}

In this section, we define Validated Task Coverage (VTC), our core idea for evaluating finite candidate sets. VTC asks a simple question: after $k$ attempts, how many distinct useful outcomes did a model produce? Each task defines what counts as useful and when two outputs count as the same.
For example, in molecule design, the goal may be to generate several chemically distinct molecules satisfying the requested properties. Close variants of the same underlying structure add little new coverage, whereas a new valid structural family increases VTC.

\textbf{Task attempts.}
For a given task, let $\mathcal{X}$ denote its set of instances. An instance $x\in\mathcal{X}$
specifies the problem presented to the model together with any task-specific constraints on a valid
response. Evaluating a model--inference configuration for $k$ attempts produces a sequence of
submitted outputs $y_{1:k}=(y_1,\ldots,y_k)$, where $y_i$ denotes the complete submission on attempt
$i$.

\textbf{VTC scoring.}
We score this $y_{1:k}$ sequence in two steps:
\begin{itemize}[leftmargin=*, itemsep=0pt, topsep=0pt]
      \item \textbf{Validation and mapping} ($\tau_x$). The function $\tau_x$ checks whether generated
            outputs are valid and satisfy the prompt constraints, then maps qualifying outputs to
            task-specific useful outcomes. Results that represent the
            same useful outcome map to the same outcome. Thus, $\tau_x(y_i)$ is the set of useful outcomes
            credited to attempt $i$, and is empty when nothing earns credit.
      \item \textbf{Coverage function} ($u_x$). We take the union of the useful outcomes across attempts, so
            encountering the same outcome again does not increase coverage. The function $u_x$ converts this
            set into the task's coverage score. Depending on the task, this score counts distinct useful
            outcomes or reports the fraction of a known target set covered.
\end{itemize}
\begin{equation}
      \text{\textbf{Validated Task Coverage}:}\quad
      \operatorname{VTC}_x(y_{1:k})
      =
      u_x\!\left(\bigcup_{i=1}^{k}\tau_x(y_i)\right).
      \label{eq:vtc}
\end{equation}
\textbf{Expected coverage under independent sampling.}
Equation~\ref{eq:vtc} scores any realized sequence of $k$ attempts and does not require the attempts
to be independent. We next consider the setting where attempts are sampled independently from a
fixed model--inference configuration $\pi(y\mid x)$. In this setting, repeated VTC measurements
estimate the expected useful coverage obtained after $k$ attempts and connect directly to standard
repeated-sampling evaluation such as pass@$k$. We therefore define
\begin{equation}
      \label{eq:expected-vtc}
      C_{\pi,\mathcal{X}}(k)
      =
      \frac{1}{|\mathcal{X}|}\sum_{x\in\mathcal{X}}
      \mathbb{E}_{y_{1:k}\sim\pi(\cdot\mid x)^k}
      \left[
            \operatorname{VTC}_x(y_{1:k})
            \right].
\end{equation}
The quantity $C_{\pi,\mathcal{X}}(k)$ is the expected coverage at attempt budget $k$ under
independent sampling. More generally, evaluating VTC across values of $k$ gives a coverage curve,
whether attempts are independent or generated sequentially. The curve increases when additional
attempts produce useful outcomes not seen earlier and flattens when they repeat previous outcomes.
Configurations with similar coverage at one attempt can therefore have different coverage at
larger $k$.

\textbf{Coverage estimation under independent sampling.}
Independent sampling also gives a convenient estimator of expected coverage. Given a larger pool
of independent attempts, we can average VTC over all size-$k$ subsets rather than relying on a
single set of $k$ draws. This yields the subset-counting estimator described in
Appendix~\ref{app:sampling}, analogous to the standard pass@$k$ estimator. Pass@$k$ is recovered
when every successful attempt maps to the same single useful outcome and every unsuccessful attempt
maps to the empty set.

\section{VTC-Bench: Benchmark Suite}

We build five candidate-generation tasks across different domains, focusing on settings where repeated outputs are genuinely useful while retaining fully automatic and reproducible evaluation. Each task is grounded in real data, provides an automatic quality check, and maps valid outputs deterministically to task-relevant useful outcomes. Together, these properties let us evaluate useful coverage directly
with task-grounded scoring, without relying on model-based judges. Table~\ref{tab:tasks} summarizes each task's attempt format, validation rule, one-draw quality measure, useful-outcome definition, and VTC calculation. Appendix~\ref{app:task-details} discusses further the considered design constraints.

\textbf{Molecule design.}
\begin{itemize}[leftmargin=*, itemsep=1pt, topsep=1pt]
      \item \textit{Why repeated candidates matter.} In molecule design, a model is asked to propose molecules that satisfy specified physicochemical constraints. Many structurally different molecules can satisfy the same constraints, so additional attempts are useful when they explore multiple on-spec structural cores rather than returning close variants of one.
      \item \textit{Construction and attempt.} Following prior work on open molecule generation with LLMs, including TOMG-Bench \citep{li2024tomg}, we construct 164 prompts that specify one or two physicochemical property ranges. The ranges are derived from ZINC250k \citep{gomezbombarelli2018automatic} so that each prompt defines a region of molecular space with many known feasible structures. On each attempt, the model returns one molecule as a SMILES string.

      \item \textit{Validation and credit.} RDKit \citep{rdkit2026} parses and sanitizes each generated molecule, checks whether it satisfies the requested property ranges, and extracts its Bemis--Murcko scaffold, a representation of the molecule's core ring-and-linker structure. Only on-spec molecules contribute to coverage, and molecules with the same scaffold count as the same useful outcome. VTC at budget $k$ counts the distinct on-spec scaffolds found across attempts.
\end{itemize}

\textbf{Repository repair.}
\begin{itemize}[leftmargin=*, itemsep=1pt, topsep=0pt]
      \item \textit{Why repeated candidates matter.} When repairing a software vulnerability, there may be several valid ways to modify the repository, targeting different parts of the code or using different implementation strategies. Repeated generation can therefore provide several repair options that an engineer may want to compare before choosing one.
      \item \textit{Construction and attempt.} We use all 230 runnable instances from PatchEval-Verified, spanning Python, JavaScript, and Go repositories \citep{wei2025patcheval}. On each attempt, the model receives a fresh copy of the full repository, can inspect and edit the source but cannot run the project or tests, and submits one final Git diff.
      \item \textit{Validation and credit.} Each patch is applied and tested for the target vulnerability in PatchEval-Verified's isolated environment. Passing patches are grouped by the set of named functions they modify, so patches that intervene in the same parts of the code count as the same useful outcome. VTC at budget $k$ counts the distinct passing modified-function sets found across attempts.
\end{itemize}

\textbf{Bug finding.}
\begin{itemize}[leftmargin=*, itemsep=1pt, topsep=1pt]
      \item \textit{Why repeated candidates matter.} In bug finding, the model generates test inputs intended to expose failures in incorrect implementations of a programming problem. A single generated suite may reveal only some of these failures, while additional suites can uncover different faulty behaviors that were missed before.
      \item \textit{Construction and attempt.} We construct 188 tasks from CodeContests C++ problems with verified accepted and incorrect submissions \citep{li2022competition}. Incorrect submissions are grouped into behavioral bug classes according to which benchmark tests they fail, giving 6,422 classes in total. On each attempt, the model receives the problem statement and returns up to five complete standard-input tests.
      \item \textit{Validation and credit.} A generated input is valid when the verified reference
            solutions agree on its output. Each valid input is executed against the incorrect submissions
            and credited with every behavioral bug class it exposes; an attempt contributes the union over
            its inputs. VTC at budget $k$ is the fraction of the task's fixed bug classes exposed across
            attempts.
\end{itemize}

\textbf{Differential diagnosis.}
\begin{itemize}[leftmargin=*, itemsep=1pt, topsep=1pt]
      \item \textit{Why repeated candidates matter.} A differential diagnosis lists several
            plausible explanations for a case. Additional attempts should broaden this list with clinically
            distinct possibilities, not merely rephrase the same diagnoses.
      \item \textit{Construction and attempt.} We use 302 NEJM clinicopathological-conference cases from the AMIE differential-diagnosis study \citep{mcduff2025accurate}. Each case provides a reconstructed History of Present Illness and a human-written reference differential. We map the reference diagnoses to a fixed set of UMLS concepts. On each attempt, the model produces a ranked differential of at most seven diagnoses and maps them to UMLS Concept Unique Identifiers (CUIs) using frozen local search tools, without access to the reference concepts.
      \item \textit{Validation and credit.} Submitted CUIs are checked for existence in the frozen
            index and provenance in the model's recorded search. Credit is restricted to submitted CUIs in
            the case's fixed reference concepts. Diagnosis-concept precision is the fraction of submitted
            concepts in that target; diagnosis-concept recall at budget $k$ is the fraction of the target
            recovered across attempts.
\end{itemize}

\textbf{BRIGHT-Pro evidence search.}
\begin{itemize}[leftmargin=*, itemsep=1pt, topsep=1pt]
      \item \textit{Why repeated candidates matter.} Answering a complex question often requires
            several pieces of evidence, each supporting a different part of the answer. Additional search
            attempts should cover these different needs rather than return several passages containing the same
            information.
      \item \textit{Construction and attempt.} We use BRIGHT-Pro \citep{zhao2026brightpro}, where complex questions are annotated with distinct reasoning aspects and supporting gold passages for each aspect. These annotations let us distinguish evidence that supports different parts of an answer from redundant evidence for the same part. On each attempt, the model issues one to three search queries against a frozen domain-specific BM25 index, inspects up to fifteen retrieved passages, and selects one to five passages as evidence.
      \item \textit{Validation and credit.} Selected passages receive credit when they are annotated gold passages, and each such passage contributes the reasoning aspect it supports. Multiple passages supporting the same aspect therefore do not add new coverage. VTC aggregates the expert-weighted reasoning aspects covered across attempts.
\end{itemize}

\begin{table}[h]
      \caption{\textbf{VTC-Bench task definitions.} $n$ is the number of task instances. Each task defines the attempt format, one-draw quality measure, useful outcomes, and headline VTC budget $H$.}
      \label{tab:tasks}
      \centering
      \fontsize{7.75}{8}\selectfont
      \renewcommand{\arraystretch}{0.5}
      \setlength{\tabcolsep}{2pt}
      \begin{tabularx}{\linewidth}{@{}
      >{\raggedright\arraybackslash}p{0.15\linewidth}
      >{\raggedright\arraybackslash}p{0.175\linewidth}
      >{\raggedright\arraybackslash}p{0.23\linewidth}
      >{\raggedright\arraybackslash}p{0.195\linewidth}
      >{\raggedright\arraybackslash}X@{}}
      \toprule
      \textbf{Task}                   &
      \textbf{Attempt}                &
      \textbf{Validation and quality} &
      \textbf{Useful outcome(s)}      &
      \textbf{Set utility (VTC)}                                                                                                                                                         \\
      \midrule
      {\textbf{Molecule design}\newline $n=164,$\newline $H=20$}
                                      & Generate one molecule as a SMILES string.
                                      & RDKit parses and sanitizes it; quality asks whether it satisfies every requested property range.
                                      & If the molecule is on spec, credit its Bemis--Murcko scaffold (structural core).
                                      & Distinct on-spec scaffolds accumulated.                                                                                                          \\
      \addlinespace
      {\textbf{Repository repair}\newline $n=230,$\newline $H=10$}
                                      & Edit a full repository and submit one Git diff.
                                      & The diff must apply and pass the isolated test for the target vulnerability.
                                      & Modified-function set of a passing patch.
                                      & Distinct passing modified-function sets accumulated.                                                                                             \\
      \addlinespace
      {\textbf{Bug finding}\newline $n=188,$\newline $H=5$}
                                      & Propose up to five test inputs for a programming problem.
                                      & Reference solutions must agree on an input's output; useful-test rate is the fraction of generated inputs that are valid and expose a bug class.
                                      & Credit every behavioral bug class exposed by a valid input; take the union within the attempt.
                                      & Fraction of fixed bug classes exposed.                                                                                                           \\
      \addlinespace
      {\textbf{Diagnosis}\newline $n=302,$\newline $H=5$}
                                      & Produce up to seven ranked diagnoses and ground them to UMLS concepts.
                                      & Concept existence is checked; quality is diagnosis-concept precision against the fixed reference target.
                                      & Credit the submitted diagnosis concepts that occur in that reference target.
                                      & Diagnosis-concept recall.                                                                                                                        \\
      \addlinespace
      {\textbf{Evidence search}\newline $n=739,$\newline $H=10$}
                                      & Issue one to three queries, inspect at most fifteen passages, and select one to five.
                                      & Search and selection calls must be well formed; quality is precision against released gold passages.
                                      & Credit the reasoning aspects supported by selected gold passages.
                                      & Reasoning-aspect recall.                                                                                                                         \\
      \bottomrule
\end{tabularx}

\end{table}

\section{Experimental Protocol}
\label{sec:protocol}

\paragraph{Generation configuration.}
We evaluate four models on all five tasks: Qwen3.6-27B, Qwen3.6-35B-A3B, GLM-4.7, and
DeepSeek-V4-Flash. For each model, we cross temperatures
$T \in \{0.6, 1.0, 1.2\}$ with thinking disabled or enabled, yielding 24 model--inference
configurations. Appendix~\ref{app:reproducibility} provides checkpoint names and generation details.

\paragraph{Repeated sampling and budgets.}
For the main 24-configuration study, we use independent repeated sampling from each
model--inference configuration, without conditioning later attempts on earlier outputs. This provides
a simple parallel reference setting with a fixed generation distribution. The task-specific $H$ values
in Table~\ref{tab:tasks} were chosen to keep candidate sets practically inspectable while controlling
generation and validation cost.  In Section~\ref{sec:chained}, we compare this reference setting with chained generation, conditioning later attempts on earlier outputs while explicitly prompting for complementary candidates. Further experimental details are presented in Appendix~\ref{app:reproducibility}.
\section{Results}
\label{sec:results}
We first characterize validated coverage across configurations and attempt budgets, then ask whether
conventional quality and surface-diversity evaluation identifies high-VTC systems. Finally, we
examine how inference choices affect quality and coverage, including temperature, thinking, and
chained generation, which conditions later attempts on earlier outputs. Results are reported for the same 24 configurations on every task, with chained generation evaluated
as a separate matched comparison. We use a task-specific headline budget $H$ for the primary comparison;
Table~\ref{tab:tasks} lists the value for each task.

\subsection{Coverage across attempt budgets}\label{sec:results-coverage}

\begin{figure}[t]
      \centering
      \begin{subfigure}[t]{0.33\linewidth}
            \centering
            \includegraphics[width=\linewidth]{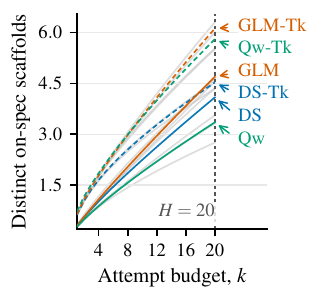}
            \caption{Molecule design.}
            \label{fig:quality-vtc-curves-molgen}
      \end{subfigure}\hfill
      \begin{subfigure}[t]{0.33\linewidth}
            \centering
            \includegraphics[width=\linewidth]{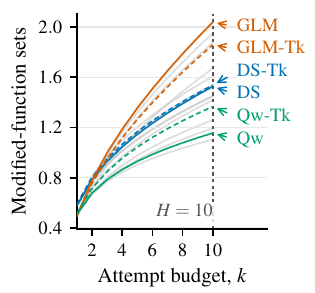}
            \caption{Repository repair.}
            \label{fig:quality-vtc-curves-patcheval}
      \end{subfigure}\hfill
      \begin{subfigure}[t]{0.33\linewidth}
            \centering
            \includegraphics[width=\linewidth]{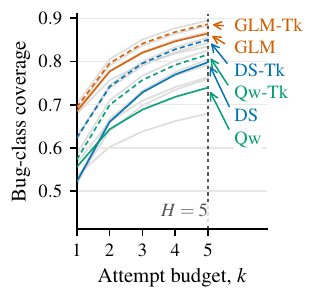}
            \caption{Bug finding.}
            \label{fig:quality-vtc-curves-bugfind}
      \end{subfigure}\par\medskip
      \begin{subfigure}[t]{0.33\linewidth}
            \centering
            \includegraphics[width=\linewidth]{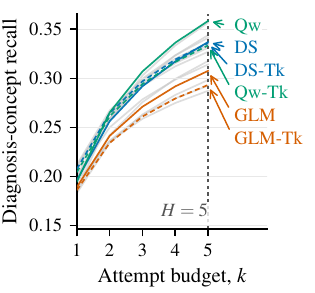}
            \caption{Differential diagnosis.}
            \label{fig:quality-vtc-curves-amie}
      \end{subfigure}\hfill
      \begin{subfigure}[t]{0.33\linewidth}
            \centering
            \includegraphics[width=\linewidth]{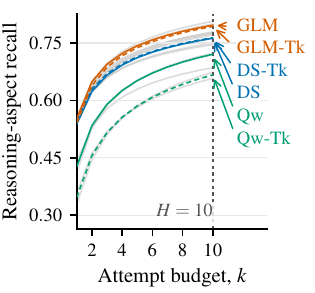}
            \caption{Evidence search.}
            \label{fig:quality-vtc-curves-bright-pro}
      \end{subfigure}\hfill
      \begin{subfigure}[t]{0.33\linewidth}
            \centering
            \includegraphics[width=\linewidth]{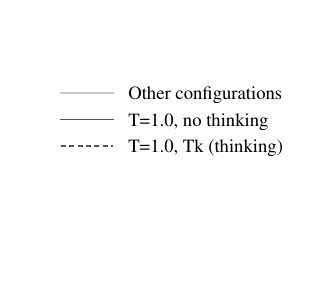}
            \caption{Plot legend.}
            \label{fig:quality-vtc-curves-legend}
      \end{subfigure}
      \caption{Each panel shows VTC evolution over the headline budget range.
            Configurations follow different coverage trajectories, producing both rank
            reversals and changing performance gaps as $k$ increases. Qwen3.6-35B-A3B is omitted for legibility.}
      \label{fig:quality-vtc-curves}
      \vspace{-0.48cm}
\end{figure}

\textbf{Coverage accumulates at different rates.} Figure~\ref{fig:quality-vtc-curves} shows how VTC exposes distinct coverage profiles across models and tasks.
GLM-4.7 configurations tend to achieve high VTC on four of the five tasks, while DeepSeek-V4-Flash is
generally competitive without dominating and Qwen3.6-27B shows more task-dependent performance.
These patterns are themselves budget-dependent: some configurations gain most of their
coverage at small $k$, while others continue to add useful outcomes, causing performance gaps to
widen or narrow as the budget increases.

\textbf{Coverage rankings change with the attempt budget.} In four of the five tasks, the configuration with the highest VTC at $k=1$ is not the one with the highest VTC at the headline budget $H$. The preferred configuration can therefore change as the attempt budget increases.
In molecule design, Qwen3.6-27B leads at $k=1$, but by $H=20$ GLM-4.7 leads by
0.502 distinct on-spec scaffolds per instance. Repository repair shows a similar pattern:
DeepSeek-V4-Flash leads at $k=1$, whereas by $H=10$, GLM-4.7 leads by 0.504 distinct
passing modified-function sets per instance. Differential diagnosis and evidence search also
change leaders, while bug finding retains the same overall leader despite crossings among other
configurations. Thus, the preferred configuration depends on the intended attempt budget, not
only on performance at a single generation. Appendix~\ref{app:vtc-budget} summarizes the consequences
of selection and the VTC changes across attempt budgets. All four leader changes remain under paired
instance resampling; details in Appendix~\ref{app:headline-robustness}.

\subsection{Quality, diversity, and coverage}\label{sec:results-selection}

Models and inference configurations are commonly compared using individual-output quality or
generic measures of variation across multiple outputs. We ask whether these signals identify the
same high-VTC configurations at a headline budget $H$. To quantify this, we rank configurations
by a selection metric and report the mean relative VTC regret of its Top-5 configurations, as
summarized in Table~\ref{tab:metric-vtc-selection}. The regret of configuration $\pi$ is
\(
\frac{C^*(H)-C_\pi(H)}{C^*(H)} \), with \(
C^*(H)=\max_\pi C_\pi(H).
\)

\begin{figure}[t]
      \centering
      \begin{subfigure}[t]{0.33\linewidth}
            \centering
            \includegraphics[width=\linewidth]{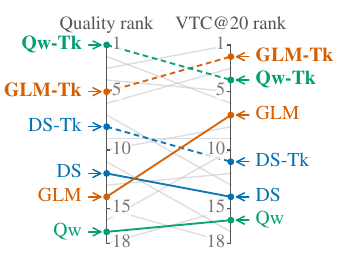}
            \caption{Molecule design.}
            \label{fig:quality-vtc-ranks-molgen}
      \end{subfigure}\hfill
      \begin{subfigure}[t]{0.33\linewidth}
            \centering
            \includegraphics[width=\linewidth]{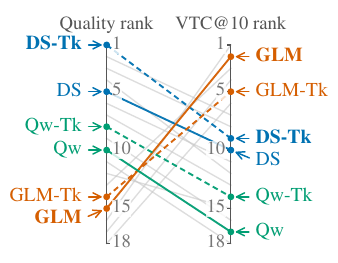}
            \caption{Repository repair.}
            \label{fig:quality-vtc-ranks-patcheval}
      \end{subfigure}\hfill
      \begin{subfigure}[t]{0.33\linewidth}
            \centering
            \includegraphics[width=\linewidth]{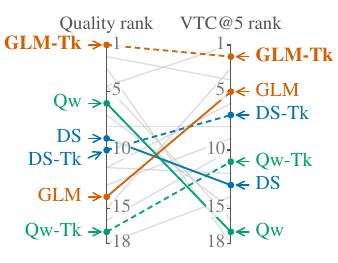}
            \caption{Bug finding.}
            \label{fig:quality-vtc-ranks-bugfind}
      \end{subfigure}\par\smallskip
      \begin{subfigure}[t]{0.33\linewidth}
            \centering
            \includegraphics[width=\linewidth]{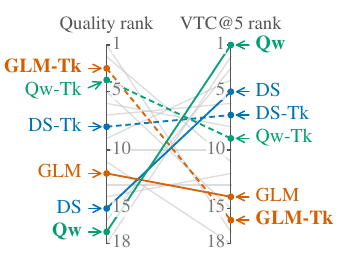}
            \caption{Differential diagnosis.}
            \label{fig:quality-vtc-ranks-amie}
      \end{subfigure}\hfill
      \begin{subfigure}[t]{0.33\linewidth}
            \centering
            \includegraphics[width=\linewidth]{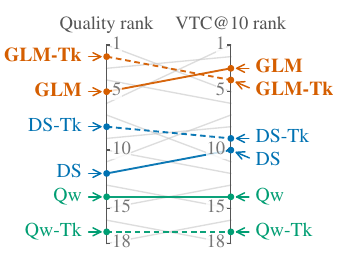}
            \caption{Evidence search.}
            \label{fig:quality-vtc-ranks-bright-pro}
      \end{subfigure}\hfill
      \begin{subfigure}[t]{0.33\linewidth}
            \centering
            \includegraphics[width=\linewidth]{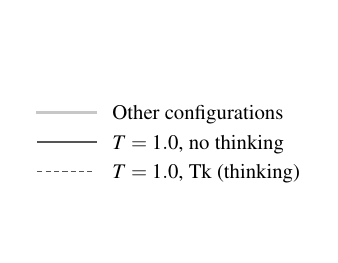}
            \caption{Plot legend.}
            \label{fig:quality-vtc-ranks-legend}
      \end{subfigure}
      \caption{\textbf{Quality and VTC rank differently.}
            Each panel compares the displayed configurations ranked by one-draw quality (left) and VTC at
            the headline budget $H$ (right). The best $T=1.0$ configuration under
            either metric is bolded.  Lines show how configurations move between
            the two rankings. Qwen3.6-35B-A3B is omitted for legibility.}
      \label{fig:quality-vtc-ranks}
\end{figure}

\textbf{One-draw quality does not reliably identify high-VTC configurations.}
Models are predominantly compared and selected using quality-focused benchmarks, so a model judged
``stronger'' is typically one that performs better on individual attempts. Figure~\ref{fig:quality-vtc-ranks} shows that these rankings can differ
substantially from rankings by VTC at the headline budget $H$. The disagreement extends beyond
the top-ranked configuration: configurations move throughout the ranking in every task. The five
configurations ranked highest by one-draw quality incur mean VTC regret ranging from 2.2\% in
evidence search to 25.0\% in repository repair, with a 12.1\% task-macro average, and share only
10 of 25 positions with the VTC Top-5 sets. Thus, high one-draw quality does not reliably identify
the configurations with the highest finite-budget VTC.

\textbf{More distinct outputs do not necessarily mean more useful coverage.}
A common way to evaluate repeated generation is to measure variation among the generated outputs
themselves, using lightweight metrics such as exact match or textual similarity. Such measures
capture surface-level variation, but differences in form need not correspond to distinct useful
outcomes. VTC instead maps validated outputs to task-specific useful outcomes. We therefore compare VTC with a budget-matched surface-diversity measure for each task, defined at a finer output level than the corresponding useful-outcome mapping (e.g., molecules rather than scaffolds, or test inputs rather than bug classes). Appendix~\ref{app:surface-diversity} defines the task-specific measures and reports the complete results.

Surface diversity does not consistently identify the configurations with the highest VTC either.
As shown in Table~\ref{tab:metric-vtc-selection}, across the five tasks, surface-diversity selection has 11.0\% task-macro mean VTC regret and
11/25 total Top-5 agreement with VTC. The mismatch is task-dependent: mean regret ranges from 3.2\% in bug finding to
25.0\% in repository repair. Thus, directly
measuring non-repetition is not sufficient to determine whether repeated outputs contribute new
useful outcomes.
The selection mismatch is stable under paired instance resampling
(Appendix~\ref{app:headline-robustness}).

\begin{table}[b]
      \caption{
            \textbf{Quality and surface diversity can misselect for VTC.}
            Top-5 mean regret is the relative VTC@$H$ shortfall of the five configurations ranked highest by each metric; agreement is their overlap with the VTC Top-5.
            Surface diversity is defined at a finer output level than VTC.}
      \label{tab:metric-vtc-selection}
      \centering
      \begingroup
      \renewcommand{\arraystretch}{1.08}
      \setlength{\tabcolsep}{4pt}
      \resizebox{0.75\linewidth}{!}{%
\begin{tabular}{@{}lrrrr@{}}
\toprule
& \multicolumn{2}{c}{\textbf{One-draw quality}} & \multicolumn{2}{c}{\textbf{Surface diversity}} \\
\cmidrule(lr){2-3}\cmidrule(l){4-5}
\textbf{Task} & \textbf{Mean regret} & \textbf{Agreement} & \textbf{Mean regret} & \textbf{Agreement} \\
\midrule
Molecule design & 7.1\% & 4/5 & 7.1\% & 4/5 \\
Repository repair & 25.0\% & 0/5 & 25.0\% & 0/5 \\
Bug finding & 12.3\% & 2/5 & 3.2\% & 3/5 \\
Differential diagnosis & 13.8\% & 0/5 & 3.5\% & 4/5 \\
Evidence search & 2.2\% & 4/5 & 16.2\% & 0/5 \\
\midrule
Task macro-average / total & 12.1\% & 10/25 & 11.0\% & 11/25 \\
\bottomrule
\end{tabular}
      }
      \endgroup
\end{table}

\subsection{Effects of inference choices on VTC}\label{sec:results-mechanisms}

We next examine inference choices that may change how useful outcomes accumulate across attempts.
We consider temperature and thinking within the independent-sampling setting, then ask whether
explicitly conditioning later attempts on earlier outputs can improve coverage.

\begin{figure}[t]
      \centering
      \begin{subfigure}[t]{0.33\linewidth}
            \centering
            \includegraphics[width=\linewidth]{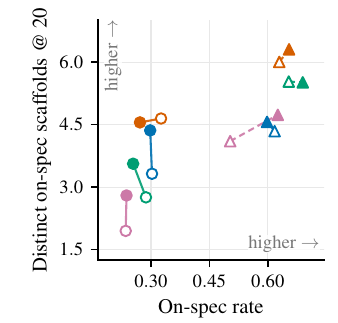}
            \caption{Molecule design.}
            \label{fig:quality-vtc-trajectories-molgen}
      \end{subfigure}\hfill
      \begin{subfigure}[t]{0.33\linewidth}
            \centering
            \includegraphics[width=\linewidth]{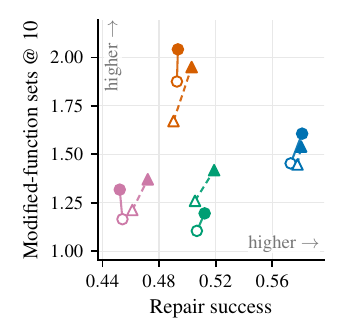}
            \caption{Repository repair.}
            \label{fig:quality-vtc-trajectories-patcheval}
      \end{subfigure}\hfill
      \begin{subfigure}[t]{0.33\linewidth}
            \centering
            \includegraphics[width=\linewidth]{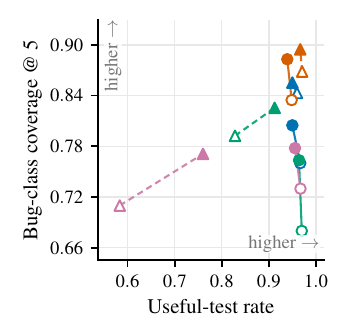}
            \caption{Bug finding.}
            \label{fig:quality-vtc-trajectories-bugfind}
      \end{subfigure}\par\smallskip
      \begin{subfigure}[t]{0.33\linewidth}
            \centering
            \includegraphics[width=\linewidth]{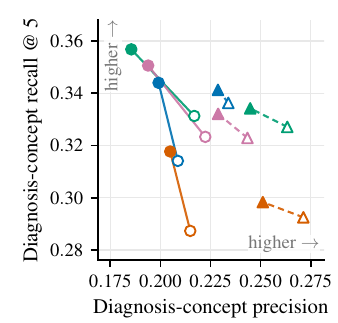}
            \caption{Differential diagnosis.}
            \label{fig:quality-vtc-trajectories-amie}
      \end{subfigure}\hfill
      \begin{subfigure}[t]{0.33\linewidth}
            \centering
            \includegraphics[width=\linewidth]{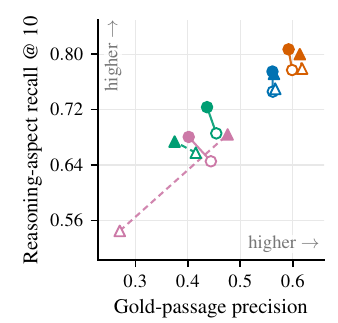}
            \caption{Evidence search.}
            \label{fig:quality-vtc-trajectories-bright-pro}
      \end{subfigure}\hfill
      \begin{subfigure}[t]{0.33\linewidth}
            \centering
            \includegraphics[width=\linewidth]{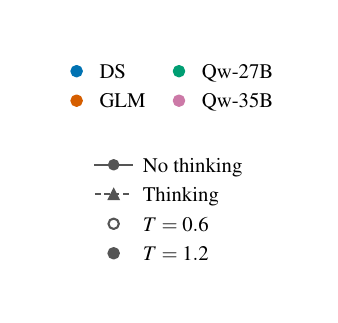}
            \caption{Plot legend.}
            \label{fig:quality-vtc-trajectories-legend}
      \end{subfigure}
      \caption{\textbf{Temperature and thinking affect quality and VTC differently.}
            Each panel places one-draw quality on the horizontal axis and VTC@$H$ on the vertical axis, with
            segments showing the effect of increasing temperature from $T=0.6$ to $T=1.2$ for matched
            model--thinking configurations. Higher temperature increases VTC@$H$ in 38 of 40 comparisons,
            often despite lower one-draw quality. Thinking introduces additional variation in both quality and
            coverage, with effects that differ across tasks.}
      \label{fig:quality-vtc-trajectories}
\end{figure}

\paragraph{Temperature can improve coverage while reducing one-draw quality.}
Figure~\ref{fig:quality-vtc-trajectories} shows that temperature can affect one-draw quality and
validated coverage in opposite directions. Increasing temperature from $0.6$ to $1.2$ raises the
point-estimate VTC@$H$ in 38 of 40 matched comparisons; in 25 of these, one-draw quality decreases.
Paired instance-resampling intervals are entirely above zero for 36 of the 40 comparisons
(Appendix~\ref{app:headline-robustness}).
Thus, an inference change that looks worse under conventional quality-focused evaluation can still
produce a more useful candidate set with higher validated coverage.

\paragraph{Thinking has heterogeneous effects on quality and coverage.}
Thinking is generally intended to improve the quality of an individual response, but this need not
translate into higher finite-budget coverage. Across the 60 matched comparisons over all three
temperatures, thinking improves one-draw quality in 44 cases but VTC@$H$ in 36. Its benefits are
therefore less consistent for coverage than for individual-output quality, and the two measures move
in opposite directions in 23 comparisons. The pattern is also heterogeneous across tasks: molecule design
generally benefits from thinking, whereas evidence search often does not.

\paragraph{Explicit diversity prompting does not reliably increase useful coverage.}\label{sec:chained}
We also evaluate a chained generation strategy explicitly designed to promote diversity: each
attempt is conditioned on earlier outputs and asked to avoid repeating previous candidates. VTC
shows that this does not reliably translate into more distinct useful outcomes. Chaining
consistently improves coverage on evidence search, consistently reduces it on repository repair,
reduces it in most differential-diagnosis configurations, has little effect on bug finding, and
shows larger configuration-dependent effects on molecule design. Thus, even when diversity is
explicitly encouraged at generation time, the resulting candidate set need not cover more validated
task-relevant outcomes. We evaluate this comparison at $H_{\mathrm{chain}}=20$ for molecule design
and $H_{\mathrm{chain}}=5$ for the other four tasks, with matched independent controls at the same
budgets. Further details and analysis appear in Appendix~\ref{app:chained-generation}.

\begin{table}[h]
      \caption{\textbf{Effect of chained generation on VTC.} Relative $\Delta$VTC compares chained
            generation with independent generation at the chained-experiment budget
            $H_{\mathrm{chain}}$ (20 for molecule design and 5 otherwise). Mean, minimum,
            and maximum changes are computed across eight matched model--inference configurations;
            ``Positive'' counts configurations for which chaining increases VTC.}
      \label{tab:chained-generation}
      \centering
      \begingroup
      \renewcommand{\arraystretch}{1.0}
      \setlength{\tabcolsep}{6pt}
      \resizebox{0.7\linewidth}{!}{%
\begin{tabular}{@{}lrrrr@{}}
\toprule
Task & Mean relative $\Delta$VTC & Min & Max & Positive \\
\midrule
Molecule design & +16.0\% & -22.0\% & +46.5\% & 6/8 \\
Repository repair & -11.5\% & -30.2\% & -0.9\% & 0/8 \\
Bug finding & +0.7\% & -2.9\% & +5.4\% & 4/8 \\
Differential diagnosis & -7.9\% & -15.4\% & +0.6\% & 1/8 \\
Evidence search & +8.3\% & +3.9\% & +13.6\% & 8/8 \\
\bottomrule
\end{tabular}
      }
      \endgroup
\end{table}

\section{Discussion and Limitations}

\textbf{VTC depends on task-specific outcome mappings.}
VTC depends on task-specific choices about which outputs are valid and which valid outputs count
as the same useful outcome. We deliberately use lightweight, deterministic mappings so that coverage
remains automatic and reproducible, but these mappings abstract away some distinctions that could
matter in practice. For example, modified-function sets capture where a patch intervenes rather than
whether two patches implement genuinely different repair strategies; diagnosis is evaluated against
a fixed UMLS target; and evidence coverage depends on the available passage and aspect annotations.
Differential diagnosis and BRIGHT-Pro additionally rely on fixed, human-curated reference targets
that may omit clinically plausible diagnoses or relevant evidence. Freezing these targets across
configurations ensures consistent comparison, but makes absolute coverage reference-relative:
valid outcomes outside the annotated target receive no credit.

\textbf{Beyond fixed attempt budgets.}
We compare configurations at a fixed number of attempts within each task, providing a simple and
controlled budget axis for repeated generation. Attempt counts are not directly comparable across
tasks in terms of token use, compute, or human review effort. The same framework could instead
measure coverage against these resource budgets when they can be tracked consistently, addressing
different practical questions from the attempt-budget analysis studied here.

\textbf{Inference choices for coverage.}
Our experiments show that inference choices can affect validated coverage in different ways: higher temperature often helps, while thinking and chained diversity prompting have more task-dependent effects. This leaves open how to choose an inference strategy that makes the best use of a fixed generation budget for a given task, a question that VTC-Bench is designed to make directly measurable.

\section{Conclusion}

VTC-Bench makes finite sets of LLM generations directly evaluable across five
tasks. By mapping validated outputs to task-relevant useful outcomes, VTC
tracks how these outcomes accumulate as the attempt budget grows. The results
show different coverage trajectories, budget-dependent rankings, and
task-dependent effects of inference choices. More broadly, our results show
that finite candidate sets exhibit measurable behavior of their own, making
them a meaningful object of model evaluation beyond individual generations.

\section*{AI use statement}
We used generative AI tools throughout the research workflow, including to assist with software
development, refine and assess the feasibility of research ideas, provide feedback on experimental
and benchmark design, and support literature search and review, particularly during the
construction of the benchmark tasks. We also used generative AI tools to assist with drafting and
editing parts of the manuscript for clarity and presentation. All AI-assisted outputs were reviewed
by the authors, and research decisions, experimental results, citations, and claims were
independently checked against the underlying code, data, and source literature. The authors take
full responsibility for the final content of the paper and all associated artifacts.

\section*{Ethics statement}
This work uses existing benchmark and published data and does not involve new data collection from
human participants. The differential-diagnosis benchmark is constructed from previously published
clinical case reports and existing AMIE-derived annotations; restricted clinical inputs and
diagnosis targets are not redistributed. The repository-repair benchmark uses real software
vulnerability instances, but evaluation is performed in isolated environments without network
access. We use these resources solely for controlled evaluation of model behavior and do not
release new sensitive clinical information or exploit code beyond the benchmark artifacts provided
by the original sources.

\section*{Reproducibility statement}
Appendix~\ref{app:reproducibility} documents the evaluated model checkpoints, inference settings,
attempt budgets, software environment, and statistical aggregation procedure.
Appendix~\ref{app:chained-generation} documents the matched chained-generation protocol and
task-specific chaining procedures, while Appendix~\ref{app:task-details} describes the construction
and deterministic scoring of each benchmark task. The later appendices report complete
configuration-level results and robustness analyses. The supplementary material provides the
frozen redistributable benchmark inputs, prompts, data provenance, and code for running the
experiments and reproducing the analyses.

\clearpage

\bibliographystyle{iclr2027_conference}
\bibliography{references}

@inproceedings{shypula2025evaluating,
  title         = {Evaluating the Diversity and Quality of {LLM} Generated Content},
  author        = {Shypula, Alexander and Li, Shuo and Zhang, Botong and
                   Padmakumar, Vishakh and Yin, Kayo and Bastani, Osbert},
  booktitle     = {Second Conference on Language Modeling},
  year          = {2025},
  eprint        = {2504.12522},
  archiveprefix = {arXiv},
  primaryclass  = {cs.CL},
  url           = {https://openreview.net/forum?id=O7bF6nlSOD}
}

@inproceedings{zhang2025noveltybench,
  title     = {{NoveltyBench}: Evaluating Language Models for Humanlike Diversity},
  author    = {Zhang, Yiming and Diddee, Harshita and Holm, Susan and Liu, Hanchen and
               Liu, Xinyue and Samuel, Vinay and Wang, Barry and Ippolito, Daphne},
  booktitle = {Second Conference on Language Modeling},
  year      = {2025},
  url       = {https://openreview.net/forum?id=XZm1ekzERf}
}

@inproceedings{ippolito2019comparison,
  title     = {Comparison of Diverse Decoding Methods from Conditional Language Models},
  author    = {Ippolito, Daphne and Kriz, Reno and Sedoc, Jo{\~a}o and
               Kustikova, Maria and Callison-Burch, Chris},
  booktitle = {Proceedings of the 57th Annual Meeting of the Association for Computational Linguistics},
  month     = jul,
  year      = {2019},
  address   = {Florence, Italy},
  publisher = {Association for Computational Linguistics},
  pages     = {3752--3762},
  doi       = {10.18653/v1/P19-1365},
  url       = {https://aclanthology.org/P19-1365/}
}

@inproceedings{alihosseini2019jointly,
  title     = {Jointly Measuring Diversity and Quality in Text Generation Models},
  author    = {Alihosseini, Danial and Montahaei, Ehsan and
               Soleymani Baghshah, Mahdieh},
  booktitle = {Proceedings of the Workshop on Methods for Optimizing and Evaluating Neural Language Generation},
  month     = jun,
  year      = {2019},
  address   = {Minneapolis, Minnesota},
  publisher = {Association for Computational Linguistics},
  pages     = {90--98},
  doi       = {10.18653/v1/W19-2311},
  url       = {https://aclanthology.org/W19-2311/}
}

@inproceedings{pillutla2021mauve,
  title     = {{MAUVE}: Measuring the Gap Between Neural Text and Human Text using Divergence Frontiers},
  author    = {Pillutla, Krishna and Swayamdipta, Swabha and Zellers, Rowan and
               Thickstun, John and Welleck, Sean and Choi, Yejin and Harchaoui, Zaid},
  booktitle = {Advances in Neural Information Processing Systems},
  volume    = {34},
  year      = {2021},
  pages     = {4816--4828},
  url       = {https://proceedings.neurips.cc/paper/2021/hash/260c2432a0eecc28ce03c10dadc078a4-Abstract.html}
}

@inproceedings{lebronnec2024exploring,
  title     = {Exploring Precision and Recall to Assess the Quality and Diversity of {LLM}s},
  author    = {Le Bronnec, Florian and Verine, Alexandre and Negrevergne, Benjamin and
               Chevaleyre, Yann and Allauzen, Alexandre},
  booktitle = {Proceedings of the 62nd Annual Meeting of the Association for Computational Linguistics (Volume 1: Long Papers)},
  month     = aug,
  year      = {2024},
  address   = {Bangkok, Thailand},
  publisher = {Association for Computational Linguistics},
  pages     = {11418--11441},
  doi       = {10.18653/v1/2024.acl-long.616},
  url       = {https://aclanthology.org/2024.acl-long.616/}
}

@article{guo2025linguistic,
  title   = {Benchmarking Linguistic Diversity of Large Language Models},
  author  = {Guo, Yanzhu and Shang, Guokan and Clavel, Chlo{\'e}},
  journal = {Transactions of the Association for Computational Linguistics},
  volume  = {13},
  year    = {2025},
  pages   = {1507--1526},
  doi     = {10.1162/tacl.a.47},
  url     = {https://aclanthology.org/2025.tacl-1.69/}
}

@article{mcduff2025accurate,
  title   = {Towards Accurate Differential Diagnosis with Large Language Models},
  author  = {McDuff, Daniel and Schaekermann, Mike and Tu, Tao and others},
  journal = {Nature},
  volume  = {642},
  year    = {2025},
  pages   = {451--457},
  doi     = {10.1038/s41586-025-08869-4},
  url     = {https://doi.org/10.1038/s41586-025-08869-4}
}

@inproceedings{zhao2026brightpro,
  title     = {Rethinking Reasoning-Intensive Retrieval: Evaluating and Advancing Retrievers in
               Agentic Search Systems},
  author    = {Zhao, Yilun and Wei, Jinbiao and Song, Tingyu and Zhang, Siyue and
               Zhao, Chen and Cohan, Arman},
  booktitle = {Proceedings of the 64th Annual Meeting of the Association for Computational
               Linguistics (Volume 1: Long Papers)},
  year      = {2026},
  pages     = {36776--36806},
  doi       = {10.18653/v1/2026.acl-long.1705},
  url       = {https://aclanthology.org/2026.acl-long.1705/}
}

@misc{chen2021evaluating,
  title         = {Evaluating Large Language Models Trained on Code},
  author        = {Chen, Mark and Tworek, Jerry and Jun, Heewoo and Yuan, Qiming and
                   Pinto, Henrique Ponde de Oliveira and Kaplan, Jared and Edwards, Harri and
                   Burda, Yuri and Joseph, Nicholas and Brockman, Greg and Ray, Alex and
                   Puri, Raul and Krueger, Gretchen and Petrov, Michael and Khlaaf, Heidy and
                   Sastry, Girish and Mishkin, Pamela and Chan, Brooke and Gray, Scott and
                   Ryder, Nick and Pavlov, Mikhail and Power, Alethea and Kaiser, Lukasz and
                   Bavarian, Mohammad and Winter, Clemens and Tillet, Philippe and
                   Such, Felipe Petroski and Cummings, Dave and Plappert, Matthias and
                   Chantzis, Fotios and Barnes, Elizabeth and Herbert-Voss, Ariel and
                   Guss, William Hebgen and Nichol, Alex and Paino, Alex and Tezak, Nikolas and
                   Tang, Jie and Babuschkin, Igor and Balaji, Suchir and Jain, Shantanu and
                   Saunders, William and Hesse, Christopher and Carr, Andrew N. and Leike, Jan and
                   Achiam, Josh and Misra, Vedant and Morikawa, Evan and Radford, Alec and
                   Knight, Matthew and Brundage, Miles and Murati, Mira and Mayer, Katie and
                   Welinder, Peter and McGrew, Bob and Amodei, Dario and McCandlish, Sam and
                   Sutskever, Ilya and Zaremba, Wojciech},
  year          = {2021},
  eprint        = {2107.03374},
  archiveprefix = {arXiv},
  primaryclass  = {cs.LG},
  url           = {https://arxiv.org/abs/2107.03374}
}

@inproceedings{wang2023selfconsistency,
  title     = {Self-Consistency Improves Chain of Thought Reasoning in Language Models},
  author    = {Wang, Xuezhi and Wei, Jason and Schuurmans, Dale and Le, Quoc V. and
               Chi, Ed H. and Narang, Sharan and Chowdhery, Aakanksha and Zhou, Denny},
  booktitle = {International Conference on Learning Representations},
  year      = {2023},
  url       = {https://openreview.net/forum?id=1PL1NIMMrw}
}

@inproceedings{huang2025bestofn,
  title     = {Is Best-of-{N} the Best of Them? Coverage, Scaling, and Optimality in
               Inference-Time Alignment},
  author    = {Huang, Audrey and Block, Adam and Liu, Qinghua and Jiang, Nan and
               Krishnamurthy, Akshay and Foster, Dylan J.},
  booktitle = {International Conference on Machine Learning},
  year      = {2025},
  url       = {https://openreview.net/forum?id=QnjfkhrbYK}
}

@inproceedings{lee2025diversely,
  title     = {How Diversely Can Language Models Solve Problems? Exploring the
               Algorithmic Diversity of Model-Generated Code},
  author    = {Lee, Seonghyeon and Chon, HeeJae and Jang, Joonwon and
               Lee, Dongha and Yu, Hwanjo},
  booktitle = {Findings of the Association for Computational Linguistics: EMNLP 2025},
  year      = {2025},
  pages     = {152--167},
  doi       = {10.18653/v1/2025.findings-emnlp.10},
  url       = {https://aclanthology.org/2025.findings-emnlp.10/}
}

@article{li2024tomg,
  title         = {{TOMG-Bench}: Evaluating {LLM}s on Text-Based Open Molecule Generation},
  author        = {Li, Jiatong and Li, Junxian and Liu, Yunqing and Zhou, Dongzhan and Li, Qing},
  journal       = {arXiv preprint arXiv:2412.14642},
  year          = {2024},
  doi           = {10.48550/arXiv.2412.14642},
  url           = {https://arxiv.org/abs/2412.14642}
}

@article{gomezbombarelli2018automatic,
  title   = {Automatic Chemical Design Using a Data-Driven Continuous Representation of Molecules},
  author  = {G{\'o}mez-Bombarelli, Rafael and Wei, Jennifer N. and Duvenaud, David and
             Hern{\'a}ndez-Lobato, Jos{\'e} Miguel and S{\'a}nchez-Lengeling, Benjam{\'i}n and
             Sheberla, Dennis and Aguilera-Iparraguirre, Jorge and Hirzel, Timothy D. and
             Adams, Ryan P. and Aspuru-Guzik, Al{\'a}n},
  journal = {ACS Central Science},
  volume  = {4},
  number  = {2},
  pages   = {268--276},
  year    = {2018},
  doi     = {10.1021/acscentsci.7b00572},
  url     = {https://doi.org/10.1021/acscentsci.7b00572}
}

@article{bemis1996properties,
  title   = {The Properties of Known Drugs. 1. Molecular Frameworks},
  author  = {Bemis, G. W. and Murcko, M. A.},
  journal = {Journal of Medicinal Chemistry},
  volume  = {39},
  number  = {15},
  pages   = {2887--2893},
  year    = {1996},
  doi     = {10.1021/jm9602928},
  url     = {https://doi.org/10.1021/jm9602928}
}

@misc{rdkit2026,
  author       = {{RDKit contributors}},
  title        = {{RDKit}: Open-Source Cheminformatics},
  year         = {2026},
  howpublished = {Version 2026.03.3},
  doi          = {10.5281/zenodo.20446949},
  url          = {https://www.rdkit.org/}
}

@article{li2022competition,
  title   = {Competition-Level Code Generation with {AlphaCode}},
  author  = {Li, Yujia and others},
  journal = {Science},
  volume  = {378},
  number  = {6624},
  pages   = {1092--1097},
  year    = {2022},
  doi     = {10.1126/science.abq1158},
  url     = {https://doi.org/10.1126/science.abq1158}
}

@misc{wei2025patcheval,
  title         = {{PATCHEVAL}: A New Benchmark for Evaluating {LLM}s on Patching Real-World
                   Vulnerabilities},
  author        = {Wei, Zichao and Zeng, Jun and Wen, Ming and Yu, Zeliang and Cheng, Kai and
                   Zhu, Yiding and Guo, Jingyi and Zhou, Shiqi and Yin, Le and Su, Xiaodong and
                   Ma, Zhechao},
  year          = {2025},
  eprint        = {2511.11019},
  archiveprefix = {arXiv},
  primaryclass  = {cs.CR},
  doi           = {10.48550/arXiv.2511.11019},
  url           = {https://arxiv.org/abs/2511.11019}
}

@inproceedings{chen2026hypospace,
  title     = {HypoSpace: A Diagnostic Benchmark for Set-Valued Hypothesis Generation under Underdetermination and Sublinear Coverage Bounds},
  author    = {Tingting Chen and Beibei Lin and Zifeng Yuan and Qiran Zou and Hongyu He and Anirudh Goyal and Yew-Soon Ong and Dianbo Liu},
  booktitle = {Forty-third International Conference on Machine Learning},
  year      = {2026},
  url       = {https://openreview.net/forum?id=QpjtK65JHO}
}

@inproceedings{mundada2026pluralism,
  title     = {Evaluating Language Model Pluralism through In-the-wild Crowd Discussions},
  author    = {Mundada, Gagan and Surana, Rohan and Swaminathan, Nandhini and
               Majumder, Bodhisattwa Prasad and Wu, Junda and McAuley, Julian and Xie, Zhouhang},
  booktitle = {Proceedings of the 64th Annual Meeting of the Association for Computational
               Linguistics (Volume 1: Long Papers)},
  year      = {2026},
  pages     = {42273--42296},
  doi       = {10.18653/v1/2026.acl-long.1957},
  url       = {https://aclanthology.org/2026.acl-long.1957/}
}

@inproceedings{chen2014systematic,
  title     = {A Systematic Comparison of Smoothing Techniques for Sentence-Level {BLEU}},
  author    = {Chen, Boxing and Cherry, Colin},
  booktitle = {Proceedings of the Ninth Workshop on Statistical Machine Translation},
  year      = {2014},
  pages     = {362--367},
  publisher = {Association for Computational Linguistics},
  doi       = {10.3115/v1/W14-3346},
  url       = {https://aclanthology.org/W14-3346/}
}

\clearpage
\appendix

\section{Experimental and Reproducibility Details}
\label{app:reproducibility}

\subsection{Evaluated generation settings}

Table~\ref{tab:generation-settings} records the settings used for the five tasks.
Every draw uses $p=0.95$ and one of the three temperatures
in the main study. The evaluated checkpoints are
Qwen3.6-27B-FP8, Qwen3.6-35B-A3B-FP8, GLM-4.7-FP8, and DeepSeek-V4-Flash. Thinking is controlled
through each model family's chat template, with the checkpoint and task prompt held fixed. The
output cap includes hidden reasoning when thinking is enabled. PatchEval uses the same cap in both
modes because its agent trajectory contains multiple model calls. In differential diagnosis, the
generation and concept-coding stages use the same temperature, thinking setting, and output cap.
In BRIGHT-Pro, the search and selection turns use the same sampled configuration and output cap.

\begin{table}[h]
      \caption{\textbf{Task-level generation settings.} $H$ is the headline comparison budget;
            complete curves are retained through $K_{\max}$.}
      \label{tab:generation-settings}
      \centering
      \small
      \setlength{\tabcolsep}{5pt}
      \begin{tabular}{@{}lrrrrl@{}}
            \toprule
            Task                   & Instances & $K_{\max}$ & $H$ & No-think / think cap & Additional limit      \\
            \midrule
            Molecule               & 164       & 50         & 20  & 256 / 32,768         & one SMILES            \\
            Repository repair      & 230       & 20         & 10  & 16,384 / 16,384      & 120 agent messages    \\
            Bug finding            & 188       & 20         & 5   & 8,192 / 32,768       & at most 5 inputs      \\
            Differential diagnosis & 302       & 20         & 5   & 2,048 / 32,768       & 7 diagnoses           \\
            BRIGHT-Pro             & 739       & 20         & 10  & 2,048 / 32,768       & 3 queries; 5 passages \\
            \bottomrule
      \end{tabular}
\end{table}

\subsection{Independent-sampling estimator and statistical aggregation}
\label{app:sampling}

For each task result, all unreduced draw-level scores are grouped by stable task-instance
identifiers before computing coverage curves, single-draw diagnostics, and diversity statistics.
Every draw enters aggregation regardless of its outcome. The analysis restricts each task to the
same set of 24 configurations and applies the task-specific budget $H$ in
Table~\ref{tab:generation-settings}. The same aggregation procedure is applied independently within
each task.

We use the subset-counting construction of the unbiased pass@$k$ estimator introduced in the Codex
evaluation of \citet{chen2021evaluating}, applying it separately to each useful outcome and summing
the resulting inclusion probabilities with task-specific weights. Under IID sampling from a fixed
configuration $\pi(\cdot\mid x)$, this gives an unbiased estimator of the expected VTC obtained from
$k$ fresh draws for instance $x$. For an instance $x$ with $K_x$
scored attempts, let $n_{x,t}$ be the number of attempts containing useful outcome $t$, and let
$w_{x,t}$ be that outcome's task-specific utility weight. For every $k\leq K_x$, we compute
\begin{equation}
      \label{eq:empirical-coverage}
      \widehat C_x(k)
      =\sum_{t\in\mathcal T_x} w_{x,t}
      \left[1-\frac{\binom{K_x-n_{x,t}}{k}}{\binom{K_x}{k}}\right].
\end{equation}
The bracketed term is the fraction of size-$k$ subsets of the observed attempts that contain useful
outcome $t$, so Equation~\ref{eq:empirical-coverage} is the exact average coverage over all such
subsets. We average $\widehat C_x(k)$ equally across the task instances. As with pass@$k$ estimation, collecting $K_{\max}>H$ reduces the sampling variance of the estimated coverage for $k\leq H$.

\subsection{Software, environment, and data provenance}

\paragraph{Software and environment.}
The runs use Inspect AI 0.3.241, with models served locally using vLLM 0.21.0. The evaluated
checkpoints were obtained from the public Hugging Face repositories
\texttt{Qwen/Qwen3.6-27B-FP8}, \texttt{Qwen/Qwen3.6-35B-A3B-FP8},
\texttt{zai-org/GLM-4.7-FP8}, and \texttt{deepseek-ai/DeepSeek-V4-Flash}.
Generation is stochastic, so independent reruns draw new completions from the same sampling
distribution. Surface SelfBLEU is computed with NLTK 3.10.0.

\paragraph{Data availability.}
Redistributable frozen benchmark inputs and prompt templates accompany the paper. Upstream release
or revision information is specified when available. The restricted AMIE HPI input,
diagnosis-concept target, and term-level coding records are not redistributed.

\section{Chained Generation Experiments}
\label{app:chained-generation}

\subsection{Overall setting and analysis}

\paragraph{Chained generation setting.}
We compare chained generation against a matched IID baseline with the same total number of generated
candidates per instance. The chained experiment uses $H_{\mathrm{chain}}=20$ for molecule design and
$H_{\mathrm{chain}}=5$ for the
other four tasks. Repository repair and evidence search therefore use a smaller budget than in the
main study to keep the additional chained generation and evaluation tractable. For each
instance, we generate $R=3$ independent chains and average VTC over their length-$k$ prefixes:
\begin{equation}
      \label{eq:chain-coverage}
      C_{\mathrm{chain},\mathcal{X}}(k)
      =
      \frac{1}{|\mathcal{X}|R}
      \sum_{x\in\mathcal{X}}\sum_{r=1}^{R}
      \operatorname{VTC}_x\!\left(y_{1:k}^{(r)}\right),
      \qquad R=3.
\end{equation}
Because later attempts condition on earlier outputs, arbitrary subsets of a chain are not valid
length-$k$ chained runs; we therefore score the actual prefix of each chain. The matched IID
baseline consists of $H_{\mathrm{chain}}\times R$ independent attempts per instance and is evaluated with the IID subset
estimator from Appendix~\ref{app:sampling}. Because it uses a separately generated matched control
pool, its estimates may differ slightly from the corresponding main-study results.

\paragraph{Chained generation analysis.}
We further examine whether the effect of chaining on VTC is reflected by conventional per-attempt
quality or surface-level diversity. For each model--thinking configuration at $T=1.0$, we use the
same task-specific quality and surface-diversity measures as in the main analysis. BRIGHT-Pro is
included in the VTC and quality comparisons but omitted from the surface-diversity comparison
because its paper surface proxy requires a separate retrieval-conditioned answer-generation stage,
which was not run for the chained campaign.
Tables~\ref{tab:chained-vtc-comparison}--\ref{tab:chained-surface-comparison} report VTC,
per-attempt quality, and surface diversity, respectively.

\begin{table}[h]
      \caption{\textbf{VTC under chained and independent generation.} Chain reports VTC from the
            actual length-$H_{\mathrm{chain}}$ chain prefixes; IID reports the subset-counting expectation at the
            same budget from the matched independent controls. Darker shading indicates higher
            values within each column; bold marks the largest value.}
      \label{tab:chained-vtc-comparison}
      \centering
      \scriptsize
      \setlength{\tabcolsep}{3pt}
\resizebox{\linewidth}{!}{%
    \begin{tabular}{@{}lcc*{5}{rr}@{}}
        \toprule
        \multicolumn{3}{c}{\textbf{Configuration}} & \multicolumn{2}{c}{\textbf{Molecule design}} & \multicolumn{2}{c}{\textbf{Repository repair}} & \multicolumn{2}{c}{\textbf{Bug finding}}       & \multicolumn{2}{c}{\textbf{Diagnosis}}         & \multicolumn{2}{c}{\textbf{Evidence search}}                                                                                                                                                                                                                                                                                                                                                          \\
        \cmidrule(lr){1-3} \cmidrule(lr){4-5} \cmidrule(lr){6-7} \cmidrule(lr){8-9} \cmidrule(lr){10-11} \cmidrule(lr){12-13}
        \textbf{Model}                             & \textbf{Thinking}                            & $T$                                            & Chain                                          & IID                                            & Chain                                          & IID                                            & Chain                                          & IID                                            & Chain                                          & IID                                            & Chain                                          & IID                                            \\
        \midrule
        DS-V4                                      & No                                           & 1.0                                            & \tableheat{resultpositive}{9}{4.715}           & \tableheat{resultpositive}{10}{4.130}          & \tableheat{resultpositive}{11}{1.019}          & \tableheat{resultpositive}{12}{1.183}          & \tableheat{resultpositive}{6}{0.775}           & \tableheat{resultpositive}{9}{0.794}           & \tableheat{resultpositive}{18}{\textbf{0.321}} & \tableheat{resultpositive}{13}{0.334}          & \tableheat{resultpositive}{14}{0.747}          & \tableheat{resultpositive}{15}{0.713}          \\
        \addlinespace[1.5pt]
        DS-V4                                      & Yes                                          & 1.0                                            & \tableheat{resultpositive}{13}{6.435}          & \tableheat{resultpositive}{11}{4.394}          & \tableheat{resultpositive}{15}{1.165}          & \tableheat{resultpositive}{11}{1.176}          & \tableheat{resultpositive}{12}{0.842}          & \tableheat{resultpositive}{14}{0.845}          & \tableheat{resultpositive}{12}{0.308}          & \tableheat{resultpositive}{14}{0.340}          & \tableheat{resultpositive}{18}{\textbf{0.787}} & \tableheat{resultpositive}{15}{0.712}          \\
        \addlinespace[3pt]
        GLM-4.7                                    & No                                           & 1.0                                            & \tableheat{resultpositive}{9}{4.929}           & \tableheat{resultpositive}{12}{4.753}          & \tableheat{resultpositive}{18}{\textbf{1.309}} & \tableheat{resultpositive}{18}{\textbf{1.367}} & \tableheat{resultpositive}{12}{0.839}          & \tableheat{resultpositive}{16}{0.864}          & \tableheat{resultpositive}{12}{0.307}          & \tableheat{resultpositive}{5}{0.305}           & \tableheat{resultpositive}{16}{0.766}          & \tableheat{resultpositive}{18}{0.738}          \\
        \addlinespace[1.5pt]
        GLM-4.7                                    & Yes                                          & 1.0                                            & \tableheat{resultpositive}{18}{\textbf{8.827}} & \tableheat{resultpositive}{18}{\textbf{6.047}} & \tableheat{resultpositive}{13}{1.116}          & \tableheat{resultpositive}{12}{1.205}          & \tableheat{resultpositive}{18}{\textbf{0.901}} & \tableheat{resultpositive}{18}{\textbf{0.884}} & \tableheat{resultpositive}{3}{0.286}           & \tableheat{resultpositive}{3}{0.298}           & \tableheat{resultpositive}{17}{0.773}          & \tableheat{resultpositive}{18}{\textbf{0.740}} \\
        \addlinespace[3pt]
        Qw3.6-27B                                  & No                                           & 1.0                                            & \tableheat{resultpositive}{4}{2.667}           & \tableheat{resultpositive}{7}{3.419}           & \tableheat{resultpositive}{8}{0.874}           & \tableheat{resultpositive}{3}{0.928}           & \tableheat{resultpositive}{3}{0.752}           & \tableheat{resultpositive}{3}{0.736}           & \tableheat{resultpositive}{12}{0.308}          & \tableheat{resultpositive}{18}{\textbf{0.354}} & \tableheat{resultpositive}{11}{0.724}          & \tableheat{resultpositive}{9}{0.652}           \\
        \addlinespace[1.5pt]
        Qw3.6-27B                                  & Yes                                          & 1.0                                            & \tableheat{resultpositive}{16}{7.785}          & \tableheat{resultpositive}{17}{5.863}          & \tableheat{resultpositive}{11}{1.007}          & \tableheat{resultpositive}{7}{1.035}           & \tableheat{resultpositive}{12}{0.842}          & \tableheat{resultpositive}{11}{0.812}          & \tableheat{resultpositive}{5}{0.291}           & \tableheat{resultpositive}{12}{0.333}          & \tableheat{resultpositive}{7}{0.678}           & \tableheat{resultpositive}{4}{0.597}           \\
        \addlinespace[3pt]
        Qw3.6-35B                                  & No                                           & 1.0                                            & \tableheat{resultpositive}{3}{2.148}           & \tableheat{resultpositive}{3}{2.568}           & \tableheat{resultpositive}{3}{0.690}           & \tableheat{resultpositive}{4}{0.948}           & \tableheat{resultpositive}{3}{0.747}           & \tableheat{resultpositive}{6}{0.762}           & \tableheat{resultpositive}{5}{0.291}           & \tableheat{resultpositive}{15}{0.344}          & \tableheat{resultpositive}{4}{0.648}           & \tableheat{resultpositive}{3}{0.588}           \\
        \addlinespace[1.5pt]
        Qw3.6-35B                                  & Yes                                          & 1.0                                            & \tableheat{resultpositive}{11}{5.746}          & \tableheat{resultpositive}{12}{4.657}          & \tableheat{resultpositive}{3}{0.670}           & \tableheat{resultpositive}{4}{0.959}           & \tableheat{resultpositive}{9}{0.806}           & \tableheat{resultpositive}{6}{0.765}           & \tableheat{resultpositive}{14}{0.311}          & \tableheat{resultpositive}{12}{0.331}          & \tableheat{resultpositive}{3}{0.642}           & \tableheat{resultpositive}{3}{0.592}           \\
        \bottomrule
    \end{tabular}%
}

\end{table}

The three quantities often move differently under chaining. On evidence search, chaining increases
VTC in all 8 configurations despite reducing per-attempt quality in 6/8, indicating a tradeoff
between passage precision and reasoning-aspect coverage. On bug finding, surface diversity
decreases in all 8 configurations, by 32.6\% on average, while VTC changes little overall, showing
that substantially fewer distinct concrete tests need not imply lower behavioral bug-class
coverage. Repository repair shows a different pattern: VTC decreases in all 8 configurations even
though quality and surface diversity change comparatively little. Differential diagnosis is the
clearest case where the metrics agree, with chaining generally reducing quality, surface diversity,
and VTC together.

Molecule design is more configuration-dependent. Chaining often increases both surface diversity
and VTC, particularly in several thinking configurations, but some Qwen configurations show the
opposite behavior. This variation reinforces that neither the intended purpose of the chaining
intervention nor its effect on surface-level output variation is sufficient to predict the
resulting useful coverage.

Overall, chained generation reproduces the main paper's quality--diversity--coverage mismatch under
an explicit diversity-oriented intervention: changes in per-attempt quality and output-level
variation do not reliably determine changes in validated task coverage.

\begin{table}[h]
      \caption{\textbf{Quality under chained and independent generation.} Chain is task-native
            quality averaged over all positions in the three chains; IID is quality over the
            attempt-count-matched independent control. Darker shading indicates higher values
            within each column; bold marks the largest value.}
      \label{tab:chained-quality-comparison}
      \centering
      \scriptsize
      \setlength{\tabcolsep}{3pt}
\resizebox{\linewidth}{!}{%
\begin{tabular}{@{}lcc*{5}{rr}@{}}
\toprule
\multicolumn{3}{c}{\textbf{Configuration}} & \multicolumn{2}{c}{\textbf{Molecule design}} & \multicolumn{2}{c}{\textbf{Repository repair}} & \multicolumn{2}{c}{\textbf{Bug finding}} & \multicolumn{2}{c}{\textbf{Diagnosis}} & \multicolumn{2}{c}{\textbf{Evidence search}} \\
\cmidrule(lr){1-3} \cmidrule(lr){4-5} \cmidrule(lr){6-7} \cmidrule(lr){8-9} \cmidrule(lr){10-11} \cmidrule(lr){12-13}
\textbf{Model} & \textbf{Thinking} & $T$ & Chain & IID & Chain & IID & Chain & IID & Chain & IID & Chain & IID \\
\midrule
DS-V4 & No & 1.0 & \tableheat{resultsmall}{6}{0.331} & \tableheat{resultsmall}{5}{0.302} & \tableheat{resultsmall}{15}{0.554} & \tableheat{resultsmall}{16}{0.573} & \tableheat{resultsmall}{17}{0.943} & \tableheat{resultsmall}{17}{0.957} & \tableheat{resultsmall}{3}{0.090} & \tableheat{resultsmall}{5}{0.204} & \tableheat{resultsmall}{9}{0.422} & \tableheat{resultsmall}{14}{0.560} \\
\addlinespace[1.5pt]
DS-V4 & Yes & 1.0 & \tableheat{resultsmall}{15}{0.606} & \tableheat{resultsmall}{14}{0.606} & \tableheat{resultsmall}{18}{\textbf{0.581}} & \tableheat{resultsmall}{18}{\textbf{0.596}} & \tableheat{resultsmall}{15}{0.935} & \tableheat{resultsmall}{17}{0.955} & \tableheat{resultsmall}{7}{0.104} & \tableheat{resultsmall}{11}{0.233} & \tableheat{resultsmall}{16}{0.513} & \tableheat{resultsmall}{15}{0.571} \\
\addlinespace[3pt]
GLM-4.7 & No & 1.0 & \tableheat{resultsmall}{4}{0.282} & \tableheat{resultsmall}{5}{0.294} & \tableheat{resultsmall}{3}{0.451} & \tableheat{resultsmall}{7}{0.488} & \tableheat{resultsmall}{13}{0.920} & \tableheat{resultsmall}{17}{0.943} & \tableheat{resultsmall}{14}{0.133} & \tableheat{resultsmall}{6}{0.209} & \tableheat{resultsmall}{17}{0.519} & \tableheat{resultsmall}{16}{0.599} \\
\addlinespace[1.5pt]
GLM-4.7 & Yes & 1.0 & \tableheat{resultsmall}{14}{0.582} & \tableheat{resultsmall}{15}{0.635} & \tableheat{resultsmall}{3}{0.455} & \tableheat{resultsmall}{8}{0.497} & \tableheat{resultsmall}{11}{0.912} & \tableheat{resultsmall}{16}{0.940} & \tableheat{resultsmall}{13}{0.127} & \tableheat{resultsmall}{18}{\textbf{0.264}} & \tableheat{resultsmall}{13}{0.478} & \tableheat{resultsmall}{18}{\textbf{0.622}} \\
\addlinespace[3pt]
Qw3.6-27B & No & 1.0 & \tableheat{resultsmall}{4}{0.255} & \tableheat{resultsmall}{4}{0.270} & \tableheat{resultsmall}{9}{0.502} & \tableheat{resultsmall}{9}{0.504} & \tableheat{resultsmall}{18}{\textbf{0.949}} & \tableheat{resultsmall}{18}{\textbf{0.967}} & \tableheat{resultsmall}{15}{0.135} & \tableheat{resultsmall}{3}{0.195} & \tableheat{resultsmall}{18}{\textbf{0.536}} & \tableheat{resultsmall}{6}{0.448} \\
\addlinespace[1.5pt]
Qw3.6-27B & Yes & 1.0 & \tableheat{resultsmall}{17}{0.657} & \tableheat{resultsmall}{18}{\textbf{0.726}} & \tableheat{resultsmall}{12}{0.529} & \tableheat{resultsmall}{10}{0.518} & \tableheat{resultsmall}{10}{0.904} & \tableheat{resultsmall}{13}{0.891} & \tableheat{resultsmall}{10}{0.116} & \tableheat{resultsmall}{15}{0.252} & \tableheat{resultsmall}{10}{0.437} & \tableheat{resultsmall}{3}{0.399} \\
\addlinespace[3pt]
Qw3.6-35B & No & 1.0 & \tableheat{resultsmall}{3}{0.239} & \tableheat{resultsmall}{3}{0.242} & \tableheat{resultsmall}{7}{0.482} & \tableheat{resultsmall}{3}{0.448} & \tableheat{resultsmall}{17}{0.947} & \tableheat{resultsmall}{18}{0.964} & \tableheat{resultsmall}{18}{\textbf{0.148}} & \tableheat{resultsmall}{5}{0.203} & \tableheat{resultsmall}{3}{0.344} & \tableheat{resultsmall}{4}{0.421} \\
\addlinespace[1.5pt]
Qw3.6-35B & Yes & 1.0 & \tableheat{resultsmall}{18}{\textbf{0.696}} & \tableheat{resultsmall}{14}{0.608} & \tableheat{resultsmall}{3}{0.452} & \tableheat{resultsmall}{3}{0.451} & \tableheat{resultsmall}{3}{0.867} & \tableheat{resultsmall}{3}{0.732} & \tableheat{resultsmall}{14}{0.131} & \tableheat{resultsmall}{12}{0.235} & \tableheat{resultsmall}{12}{0.461} & \tableheat{resultsmall}{8}{0.476} \\
\bottomrule
\end{tabular}%
}

\end{table}

\begin{table}[h]
      \caption{\textbf{Surface diversity under chained and independent generation.} Chain reports
            direct-prefix coverage of validated canonical molecules, accepted exact diffs, exact
            valid test inputs, and gold-mapped raw diagnosis terms; IID reports the corresponding
            subset-counting expectation at $H_{\mathrm{chain}}$. BRIGHT-Pro is omitted because its separate
            answer-generation surface experiment was not run for chained generation. Bold marks
            the largest value in each column.}
      \label{tab:chained-surface-comparison}
      \centering
      \scriptsize
      \setlength{\tabcolsep}{3pt}
      \resizebox{0.8\linewidth}{!}{%
    \begin{tabular}{@{}lcc*{4}{rr}@{}}
        \toprule
        \multicolumn{3}{c}{\textbf{Configuration}} & \multicolumn{2}{c}{\textbf{Molecule design}} & \multicolumn{2}{c}{\textbf{Repository repair}} & \multicolumn{2}{c}{\textbf{Bug finding}} & \multicolumn{2}{c}{\textbf{Diagnosis}}                                                                                                         \\
        \cmidrule(lr){1-3} \cmidrule(lr){4-5} \cmidrule(lr){6-7} \cmidrule(lr){8-9} \cmidrule(lr){10-11}
        \textbf{Model}                             & \textbf{Thinking}                            & $T$                                            & Chain                                    & IID                                    & Chain          & IID            & Chain           & IID             & Chain          & IID            \\
        \midrule
        DS-V4                                      & No                                           & 1.0                                            & 6.470                                    & 5.442                                  & 2.512          & 2.611          & 7.172           & 12.566          & \textbf{3.139} & \textbf{4.804} \\
        \addlinespace[1.5pt]
        DS-V4                                      & Yes                                          & 1.0                                            & 12.061                                   & 9.351                                  & \textbf{2.655} & \textbf{2.654} & 12.379          & 17.342          & 2.932          & 4.248          \\
        \addlinespace[3pt]
        GLM-4.7                                    & No                                           & 1.0                                            & 5.602                                    & 5.831                                  & 2.246          & 2.347          & 9.819           & 16.637          & 3.024          & 3.491          \\
        \addlinespace[1.5pt]
        GLM-4.7                                    & Yes                                          & 1.0                                            & 11.624                                   & 10.519                                 & 2.251          & 2.318          & \textbf{16.238} & \textbf{19.067} & 2.402          & 3.170          \\
        \addlinespace[3pt]
        Qw3.6-27B                                  & No                                           & 1.0                                            & 4.459                                    & 4.522                                  & 2.184          & 2.099          & 9.417           & 13.627          & 3.043          & 4.258          \\
        \addlinespace[1.5pt]
        Qw3.6-27B                                  & Yes                                          & 1.0                                            & 13.067                                   & \textbf{11.393}                        & 2.365          & 2.247          & 11.387          & 17.267          & 2.512          & 3.302          \\
        \addlinespace[3pt]
        Qw3.6-35B                                  & No                                           & 1.0                                            & 4.425                                    & 4.371                                  & 1.920          & 2.035          & 7.851           & 13.199          & 2.882          & 4.242          \\
        \addlinespace[1.5pt]
        Qw3.6-35B                                  & Yes                                          & 1.0                                            & \textbf{13.654}                          & 9.015                                  & 1.746          & 2.037          & 10.512          & 14.557          & 3.011          & 3.353          \\
        \bottomrule
    \end{tabular}%
      }
\end{table}

\subsection{Task-specific chaining procedures}

Across tasks, the first attempt uses the original task prompt. Each later attempt receives a bounded
record of the model's earlier submissions and a task-specific request for one additional candidate.
The history contains neither validation results nor correctness feedback, and each of the three
chain replicates starts without history from the others.

\paragraph{Molecule design.}
The model sees the SMILES strings submitted on earlier attempts and is asked for one additional
molecule that satisfies the original constraints while being materially different from its prior
submissions. Each attempt retains the original one-SMILES output format.

\paragraph{Repository repair.}
The model sees the patches submitted earlier in the chain, but not their validation results. The
repository is reset to the original vulnerable snapshot before every attempt, and the model is asked
to produce one materially different patch using the same editing interface as in the main study.

\paragraph{Bug finding.}
Earlier test suites are shown in the same input format used for submission. The continuation asks
for another suite based on a materially different testing idea, subject to the original limit of at
most five inputs. Outcomes from executing earlier tests are not shown to the model.

\paragraph{Differential diagnosis.}
The model sees the diagnosis terms from its earlier ranked differentials and is asked for another
substantively different but clinically plausible differential in the original list format. The UMLS
mapping process and matches against the reference differential are not included in the history.

\paragraph{Evidence search.}
The model sees the queries and selected evidence excerpts from earlier attempts and is asked to use
complementary queries and evidence for the same question. Each attempt retains the original limits
of up to three queries and five selected passages; unselected retrieval results and coverage scores
are not carried into later attempts.

\section{Benchmark Construction and Scoring}
\label{app:task-details}

\subsection{Design criteria for useful outcomes}

For each task, we choose a useful-outcome representation that can be checked automatically,
is reproducible across systems, and captures distinctions that matter for the task. Repeatedly
producing the same outcome should add little or no new coverage.

Exact output identity is often too fine-grained, while semantic grouping with embeddings or judges
would make scoring less deterministic. We therefore use task-specific intermediate representations
grounded in the structure of each task. Table~\ref{tab:unit-choices} summarizes these choices.

\begin{table}[h]
      \caption{\textbf{Task-specific useful outcomes.} Each representation abstracts away surface
            variation while remaining deterministic, automatically scoreable, and grounded in task structure.}
      \label{tab:unit-choices}
      \centering
      \scriptsize
      \renewcommand{\arraystretch}{1.16}
      \setlength{\tabcolsep}{3pt}
      \begin{tabularx}{\linewidth}{@{}
            >{\raggedright\arraybackslash}p{0.12\linewidth}
            >{\raggedright\arraybackslash}p{0.13\linewidth}
            >{\raggedright\arraybackslash}p{0.18\linewidth}
            >{\raggedright\arraybackslash}p{0.16\linewidth}
            >{\raggedright\arraybackslash}X@{}}
            \toprule
            \textbf{Task}     & \textbf{Too-fine representation} & \textbf{Requires semantic judgment}        & \textbf{Useful outcome} & \textbf{Rationale}                                                                                                                                                        \\
            \midrule
            Molecule design   & Exact molecule                   & Functional or semantic chemical similarity & Bemis--Murcko scaffold  & Collapses peripheral substituent and encoding variation while retaining the molecular structural core.                                                                    \\
            Repository repair & Exact Git diff                   & Semantic repair strategy                   & Modified-function set   & Collapses incidental diff-level variation while distinguishing repairs that intervene in different parts of the implementation.                                           \\
            Bug finding       & Exact test input                 & Underlying bug mechanism                   & Behavioral bug class    & Different concrete inputs may expose the same faulty behavior; execution against the fixed submission pool gives a deterministic behavioral grouping.                     \\
            Diagnosis         & Diagnosis string                 & Unrestricted clinical equivalence          & UMLS concept            & Collapses synonyms, abbreviations, and paraphrases that refer to the same clinical concept while retaining concept-level distinctions.                                    \\
            Evidence search   & Passage ID                       & Free-form semantic contribution            & Reasoning aspect        & Multiple passages may support the same information need; aspect annotations capture complementary pieces of evidence without requiring semantic judgment of full answers. \\
            \bottomrule
      \end{tabularx}
\end{table}

\subsection{Molecule design}

\paragraph{Choice of useful-outcome representation.}
We represent useful outcomes by Bemis--Murcko scaffolds, extracted automatically with RDKit.
Exact molecule identity is too fine-grained for our purpose: molecules can differ in peripheral
substituents while sharing the same underlying structural core, so counting them separately would
reward variation that contributes little new structural coverage. Bemis--Murcko scaffolds provide a
deterministic intermediate abstraction that retains ring systems and the linkers between them while
removing peripheral substituents \citep{bemis1996properties}. Molecules with the same scaffold
therefore map to the same useful outcome even when their SMILES encodings or peripheral
substituents differ. We use the number of distinct on-spec scaffolds accumulated across attempts as
a reproducible proxy for structural exploration under explicit physicochemical constraints.

\paragraph{Instance construction.}
We construct the benchmark from ZINC250k, a standard molecular-generation dataset of
approximately 250,000 drug-like, commercially available molecules sampled from the ZINC database
\citep{gomezbombarelli2018automatic}. We parse all entries with RDKit 2026.03.3, retaining
249,455 valid molecules. For each molecule, we record its canonical SMILES, Bemis--Murcko
scaffold, and six conditioning descriptors: molecular weight, RDKit-calculated logP, topological
polar surface area (TPSA), hydrogen-bond donors, hydrogen-bond acceptors, and rotatable bonds.
QED is excluded because it is a composite desirability score rather than a primitive property
constraint.

For each descriptor, we form the five 30-percentile-wide windows
$[Q_{.05},Q_{.35}]$, $[Q_{.20},Q_{.50}]$, $[Q_{.35},Q_{.65}]$,
$[Q_{.50},Q_{.80}]$, and $[Q_{.65},Q_{.95}]$. Adjacent windows overlap by half their width, and
their union covers the central 90\% of the marginal distribution. We round endpoints to multiples of
10 Da for molecular weight, 0.5 for logP, 5~\AA$^2$ for TPSA, and one count for the integer
descriptors. For the integer-valued descriptors, rounding can cause two quantile windows to produce the same interval. When this occurs, we keep only one copy of that interval.
We enumerate every single-descriptor band and every combination of bands over descriptor pairs. This procedure
produces 379 candidate specifications.

To avoid instances whose feasible region is too narrow in the reference pool, we require each
specification to be satisfied by at least 5,000 ZINC250k molecules and at least 1,000 distinct
Bemis--Murcko scaffolds. All 379 candidates pass this prespecified, model-independent filter,
yielding 29 one-constraint and 350 two-constraint prompts.

For the paper evaluation, we retain all 29 one-constraint specifications. For each two-descriptor
pair, we retain only combinations formed from the first, middle, and last window of each descriptor,
giving $3 \times 3 = 9$ prompts per descriptor pair. This yields 135 two-constraint prompts and
164 prompts overall.

The released prompt artifact records the exact resolved bands, constraints, support counts,
ZINC250k source identity, and RDKit version.

\paragraph{Scoring.} At scoring time, we recompute the same RDKit descriptors used during construction and apply all interval bounds inclusively. Valid on-spec molecules are then mapped to their Bemis--Murcko scaffold.

\subsection{Repository repair}

\paragraph{Benchmark setup.}
We use all 230 runnable instances from PatchEval-Verified, spanning Python, JavaScript, and Go
repositories \citep{wei2025patcheval}. Each instance provides a vulnerable repository snapshot and
an isolated environment for validating candidate patches. The task description includes the CVE
identifier and description together with its CWE category. PatchEval-Verified also provides
annotated vulnerable files and line ranges; we withhold these annotations so that the agent must
identify the relevant implementation locations from the full repository.

Each attempt begins from a fresh repository snapshot. The agent may search and read files and edit
the source, but cannot execute the project, run tests, or access the network. It returns one complete
Git diff, used as the final submission. It does not receive any validator feedback.

\paragraph{Choice of useful-outcome representation.}
We represent each successful patch by the set of named functions it modifies. Exact Git-diff
identity is too fine-grained for our purpose: two patches can differ in formatting, local edits, or
other incidental details while intervening in the same parts of the implementation. Modified-function
sets provide a deterministic intermediate representation that collapses such diff-level variation
while distinguishing successful repairs that act at different implementation locations. Patches with
the same modified-function set therefore map to the same useful outcome.

\paragraph{Evaluation details.}
A submitted patch must be a parseable Git diff that applies cleanly to the original repository. The
corresponding PatchEval-Verified environment applies the diff and runs the revised vulnerability
test with a 600-second limit. A patch earns credit only if this validation succeeds; successful
patches contribute their modified-function set, while all other attempts contribute no useful
outcome.

To construct the modified-function set, we parse each changed Python, JavaScript/TypeScript, or Go
source file and map changed lines to their innermost enclosing named function in the original
repository. Removed lines retain their original coordinates, while inserted lines are anchored to
adjacent original lines. Anonymous functions map to their nearest named enclosing function, edits
outside named functions map to a module-level site, and unparseable files remain explicit
file-level sites. The resulting sorted set of sites defines the patch's useful outcome.

\subsection{Bug finding}

\paragraph{Benchmark setup.}
We construct the task from C++ submissions in the CodeContests test and validation splits
\citep{li2022competition}. We remove problems that require file I/O or images and retain problems
with at least three accepted submissions, five incorrect submissions, and one stored test. We verify
up to five accepted submissions on a fixed probe of at most 80 stored tests and require at least
three of them to reproduce every expected output.

We retain incorrect submissions that compile and fail at least one probe test. Incorrect submissions
with the same pattern of failures on the probe tests are grouped into one behavioral bug class. This
produces 188 problems and 6,422 fixed bug classes, with a median of 30 classes per problem.

On each attempt, the model receives the problem statement and generates up to five complete
standard-input test cases. The prompt does not ask the model to rank the tests or make them distinct.

\paragraph{Choice of useful-outcome representation.}
We measure useful outcomes at the level of behavioral bug classes rather than generated test inputs.
Exact input identity is too fine-grained: different concrete inputs can expose the same faulty
behavior, so counting them separately would overstate useful coverage. Instead, each valid input is
credited with the behavioral bug classes it exposes when executed against the fixed pool of incorrect
submissions. Different inputs that expose the same bug class therefore contribute the same useful
outcome, while an input may contribute multiple outcomes if it exposes multiple classes. Credits
are unioned across the inputs in an attempt and across repeated attempts.

\paragraph{Evaluation details.}
At evaluation time, the accepted reference programs and incorrect programs are compiled under
contest-judge conditions and executed without network access, using each problem's memory limit
and a wall-time limit of $\min(t_{\mathrm{problem}}+1,\mathrm{s},6,\mathrm{s})$.

A generated input is valid when at least two reference programs finish and all finishing references
agree on the output. Outputs are compared as whitespace-separated tokens, with a numeric tolerance
of $10^{-6}$. A valid input exposes an incorrect implementation if that implementation produces a
different output, exits unsuccessfully, or exceeds a CPU or wall-time limit. The corresponding
behavioral bug class is then credited to the attempt, with each class credited at most once within an
attempt.

The scorer evaluates at most the first five parsed inputs in response order. Invalid inputs contribute
no bug-class credit but remain part of the submitted attempt.

\subsection{Differential diagnosis}

\paragraph{Benchmark setup.}
We follow the case selection of the AMIE differential-diagnosis study
\citep{mcduff2025accurate}, using its 302 NEJM clinicopathological-conference cases. For each case,
we reconstruct the History of Present Illness from the corresponding NEJM report, beginning at
``Presentation of Case'' and ending at the next clinical section, while removing tables, captions,
headers, footers, and affiliations. The reconstructed History of Present Illness is the sole clinical
case information provided to the evaluated model.

\paragraph{Reference target construction.}
We map the physician-written differential diagnoses and final diagnosis to concepts in UMLS
2025AB. Our frozen local index contains English, non-suppressed UMLS names, synonyms, semantic
types, and definitions. Queries and indexed strings are normalized for case, punctuation, and
whitespace. Retrieval prioritizes exact matches, followed by BM25 matches over query tokens, and
returns the ten highest-ranked distinct Concept Unique Identifiers (CUIs).

The reference diagnoses are mapped once to CUIs using OpenAI Codex with the frozen local search tools. These mappings are then frozen as part of the benchmark target; Codex is not used to judge evaluated outputs, whose scores are computed deterministically by exact CUI matching against this fixed target. Codex may reformulate searches and inspect retrieved names, synonyms, semantic types, and
definitions before selecting one or more retrieved CUIs, or no CUI when none is appropriate. The
mapping preserves the specificity of the written diagnosis, and multiple CUIs may be retained when
the diagnosis is ambiguous. Final diagnoses are mapped independently from the published
differential. Every selected reference CUI must exist in the frozen index and have appeared among
the retrieved candidates.

The published differentials contain 2,190 terms, of which 2,179 map to at least one CUI and 148
have multiple retained mappings. Final diagnoses are mapped separately, with 296 of 302 receiving
a CUI. The target for each case is the union of the CUIs assigned to its published differential and
final diagnosis, producing target sets of size 2--19 (median 8, mean 8.32). The same fixed target is
used for every evaluated configuration.

\paragraph{Choice of useful-outcome representation.}
We represent diagnoses by UMLS CUIs rather than their generated strings. Exact diagnosis strings
are too fine-grained: synonyms, abbreviations, and paraphrases can refer to the same clinical concept
and should not count as distinct coverage. UMLS concepts provide a fixed intermediate
representation that collapses such surface variation while retaining clinically distinct concepts.
Different generated diagnoses mapped to the same CUI therefore contribute the same useful
outcome.

\paragraph{Prediction coding and evaluation.}
On each attempt, the evaluated model generates a ranked differential of at most seven diagnoses.
The same model then maps each generated diagnosis to UMLS concepts using the frozen local search
tools using the same temperature, thinking setting, and output-token allowance as differential
generation. It retains the case and its own generated differential as context but does not receive
the reference CUIs. Concept coding is therefore part of the evaluated attempt rather than a
post-hoc evaluator-side transformation.

The scorer retains at most the first seven parsed diagnoses and compares the submitted CUIs with
the fixed reference target by exact CUI intersection. Diagnosis-concept precision is the fraction of distinct submitted CUIs contained in
the target, with zero assigned when no CUI is submitted. VTC at budget $k$ is the fraction of target
CUIs recovered across the $k$ attempts.

\subsection{BRIGHT-Pro evidence search}

\paragraph{Benchmark setup.}
We use a pinned public release of the \texttt{yale-nlp/Bright-Pro} dataset
\citep{zhao2026brightpro}. It contains 739 questions from seven StackExchange domains,
2,763 expert-annotated reasoning aspects, 5,272 released gold passages, and 526,319 corpus
documents. Each question has between one and five reasoning aspects. Raw aspect weights are in
${1,2,3}$ and are normalized to sum to one within each question.

We build a separate frozen BM25 index for each domain, so each question searches only documents
from its corresponding domain. The indices use BM25S 0.3.9 with Lucene BM25 IDF,
$k_1=0.9$, $b=0.4$, lowercase regex tokenization, Porter stemming, and the Lucene English
stopword set. These settings match the original BRIGHT BM25 parameters, although our tokenizer is
a close Python reproduction rather than the original Lucene analyzer.

\paragraph{Choice of useful-outcome representation.}
We represent useful evidence at the level of annotated reasoning aspects rather than exact passage
identity. Passage identity is too fine-grained for coverage because several passages may support the
same part of the information need. Counting those passages separately would therefore reward
redundant evidence. At the other extreme, deciding whether complete generated answers contribute
different information would require semantic judgment. BRIGHT-Pro's expert reasoning-aspect
annotations provide a fixed intermediate representation of the distinct pieces of evidence needed to
answer a question. Multiple selected passages supporting the same aspect therefore contribute the
same useful outcome.

\paragraph{Retrieval and selection procedure.}
On each attempt, the model submits one batch of one to three nonempty search queries. Each query
retrieves the top 15 passages from the corresponding domain index. Results are interleaved by
within-query rank, deduplicated by document ID, and truncated to the first 15 passages. The visible
candidate bundle is further truncated under the fixed character budget specified in the released task
artifact.

The model then selects between one and five unique passage labels from the visible pool. It sees the
passage text and query/rank provenance, but not document IDs, retrieval scores, released-gold labels,
reasoning aspects, aspect weights, qrels, or the reference answer. Missing or malformed search or
selection calls are retained as zero-scoring attempts.

\paragraph{Evaluation details.}
Exact released document IDs determine whether a selected passage is a released gold passage. Each
selected gold passage contributes its annotated reasoning aspect, and an aspect receives credit at
most once within an attempt even when multiple selected passages support it. VTC at budget $k$ is
the weighted fraction of reasoning aspects recovered across the $k$ attempts, using the normalized
aspect weights.

Because the released qrels are not exhaustive relevance judgments, unjudged passages receive no
credit.

\section{Additional Results and Robustness}

\subsection{VTC rankings across attempt budgets}
\label{app:vtc-budget}

Table~\ref{tab:vtc-budget} compares configuration rankings by VTC@1 and VTC@$H$ across all
24 configurations. We select the five configurations with the highest VTC@1 and evaluate their
VTC@$H$ using the same mean relative regret as Table~\ref{tab:metric-vtc-selection}. Agreement counts
how many configurations appear in both Top-5 sets. The final column reports the mean VTC gain from
one attempt to the headline budget, averaged across all 24 configurations in each task. These gains
remain in each task's native VTC units and are not averaged across tasks.

For molecule design and repository repair, VTC@1 equals the one-draw quality metric because each
attempt contributes at most one useful outcome. In the other three tasks, one attempt can cover a
set of useful outcomes, so VTC@1 is already a set-coverage quantity and is not identical to the
quality metric used in Table~\ref{tab:metric-vtc-selection}.

\begin{table}[H]
      \caption{\textbf{VTC rankings can depend on the attempt budget.} Top-5 mean regret is the
            average relative VTC@$H$ shortfall of the five configurations selected by VTC@1.
            Agreement is their overlap with the VTC@$H$ Top-5. Mean VTC gain is the difference
            between VTC@$H$ and VTC@1, averaged across all 24 configurations and reported on each
            task's native scale.}
      \label{tab:vtc-budget}
      \centering
\footnotesize
\renewcommand{\arraystretch}{1.08}
\setlength{\tabcolsep}{5pt}
\begin{tabular}{@{}lrrr@{}}
\toprule
\textbf{Task} & \textbf{Mean regret} & \textbf{Agreement} & \textbf{Mean VTC gain} \\
\midrule
Molecule design & 7.1\% & 4/5 & 3.919 \\
Repository repair & 25.0\% & 0/5 & 0.974 \\
Bug finding & 1.7\% & 5/5 & 0.231 \\
Differential diagnosis & 6.7\% & 1/5 & 0.131 \\
Evidence search & 4.3\% & 1/5 & 0.277 \\
\midrule
Task macro-average / total & 9.0\% & 11/25 & --- \\
\bottomrule
\end{tabular}

\end{table}
\clearpage

\subsection{Full configuration results}

Table~\ref{tab:full-configuration-results} reports both requested selection quantities for every
combination of model, thinking setting, and temperature. The task-specific definitions of one-draw
quality and VTC are given in Table~\ref{tab:tasks}; values are kept on their native task scales
rather than normalized across tasks. The complete tables abbreviate DeepSeek-V4-Flash,
Qwen3.6-27B, and Qwen3.6-35B-A3B as DS-V4, Qw3.6-27B, and Qw3.6-35B.
\begin{table}[h]
      \caption{\textbf{Quality and VTC for all 24 configurations.} Each task reports its one-draw
            quality measure and VTC at headline budget $H$. Quality is, from left to right, on-spec rate,
            repair success, useful-test rate, diagnosis-concept precision, and gold-passage precision. Darker
            shading indicates higher values within each task--metric column; bold marks the largest value.}
      \label{tab:full-configuration-results}
      \centering
      \scriptsize
      \setlength{\tabcolsep}{3pt}
\resizebox{\linewidth}{!}{%
\begin{tabular}{@{}lcc*{5}{rr}@{}}
\toprule
\multicolumn{3}{c}{\textbf{Configuration}} & \multicolumn{2}{c}{\textbf{Molecule design}} & \multicolumn{2}{c}{\textbf{Repository repair}} & \multicolumn{2}{c}{\textbf{Bug finding}} & \multicolumn{2}{c}{\textbf{Diagnosis}} & \multicolumn{2}{c}{\textbf{Evidence search}} \\
\cmidrule(lr){1-3} \cmidrule(lr){4-5} \cmidrule(lr){6-7}
\cmidrule(lr){8-9} \cmidrule(lr){10-11} \cmidrule(lr){12-13}
\textbf{Model} & \textbf{Thinking} & $T$ & Quality & VTC@20 & Quality & VTC@10 & Quality & VTC@5 & Quality & VTC@5 & Quality & VTC@10 \\
\midrule
DS-V4 & No & 0.6 & \tableheat{resultsmall}{5}{0.303} & \tableheat{resultpositive}{8}{3.322} & \tableheat{resultsmall}{16}{0.573} & \tableheat{resultpositive}{9}{1.453} & \tableheat{resultsmall}{18}{0.967} & \tableheat{resultpositive}{9}{0.760} & \tableheat{resultsmall}{7}{0.209} & \tableheat{resultpositive}{9}{0.314} & \tableheat{resultsmall}{16}{0.562} & \tableheat{resultpositive}{15}{0.746} \\
DS-V4 & No & 1.0 & \tableheat{resultsmall}{5}{0.299} & \tableheat{resultpositive}{10}{4.089} & \tableheat{resultsmall}{17}{0.576} & \tableheat{resultpositive}{10}{1.527} & \tableheat{resultsmall}{17}{0.957} & \tableheat{resultpositive}{11}{0.798} & \tableheat{resultsmall}{6}{0.205} & \tableheat{resultpositive}{13}{0.337} & \tableheat{resultsmall}{15}{0.557} & \tableheat{resultpositive}{15}{0.763} \\
DS-V4 & No & 1.2 & \tableheat{resultsmall}{5}{0.298} & \tableheat{resultpositive}{11}{4.362} & \tableheat{resultsmall}{17}{0.581} & \tableheat{resultpositive}{11}{1.607} & \tableheat{resultsmall}{17}{0.950} & \tableheat{resultpositive}{12}{0.805} & \tableheat{resultsmall}{5}{0.199} & \tableheat{resultpositive}{15}{0.344} & \tableheat{resultsmall}{16}{0.561} & \tableheat{resultpositive}{16}{0.775} \\
\addlinespace[1.5pt]
DS-V4 & Yes & 0.6 & \tableheat{resultsmall}{15}{0.618} & \tableheat{resultpositive}{11}{4.344} & \tableheat{resultsmall}{17}{0.578} & \tableheat{resultpositive}{8}{1.448} & \tableheat{resultsmall}{18}{0.960} & \tableheat{resultpositive}{14}{0.843} & \tableheat{resultsmall}{11}{0.234} & \tableheat{resultpositive}{13}{0.336} & \tableheat{resultsmall}{16}{0.567} & \tableheat{resultpositive}{15}{0.750} \\
DS-V4 & Yes & 1.0 & \tableheat{resultsmall}{15}{0.616} & \tableheat{resultpositive}{12}{4.529} & \tableheat{resultsmall}{18}{\textbf{0.588}} & \tableheat{resultpositive}{10}{1.537} & \tableheat{resultsmall}{17}{0.950} & \tableheat{resultpositive}{15}{0.850} & \tableheat{resultsmall}{11}{0.229} & \tableheat{resultpositive}{13}{0.336} & \tableheat{resultsmall}{16}{0.567} & \tableheat{resultpositive}{16}{0.764} \\
DS-V4 & Yes & 1.2 & \tableheat{resultsmall}{14}{0.598} & \tableheat{resultpositive}{12}{4.566} & \tableheat{resultsmall}{17}{0.580} & \tableheat{resultpositive}{10}{1.540} & \tableheat{resultsmall}{17}{0.950} & \tableheat{resultpositive}{15}{0.855} & \tableheat{resultsmall}{11}{0.229} & \tableheat{resultpositive}{14}{0.341} & \tableheat{resultsmall}{16}{0.564} & \tableheat{resultpositive}{16}{0.772} \\
\addlinespace[3pt]
GLM-4.7 & No & 0.6 & \tableheat{resultsmall}{6}{0.326} & \tableheat{resultpositive}{12}{4.648} & \tableheat{resultsmall}{7}{0.493} & \tableheat{resultpositive}{15}{1.875} & \tableheat{resultsmall}{17}{0.948} & \tableheat{resultpositive}{14}{0.835} & \tableheat{resultsmall}{8}{0.215} & \tableheat{resultpositive}{3}{0.287} & \tableheat{resultsmall}{17}{0.599} & \tableheat{resultpositive}{16}{0.777} \\
GLM-4.7 & No & 1.0 & \tableheat{resultsmall}{5}{0.291} & \tableheat{resultpositive}{12}{4.697} & \tableheat{resultsmall}{8}{0.494} & \tableheat{resultpositive}{18}{2.038} & \tableheat{resultsmall}{17}{0.945} & \tableheat{resultpositive}{16}{0.865} & \tableheat{resultsmall}{7}{0.210} & \tableheat{resultpositive}{7}{0.308} & \tableheat{resultsmall}{17}{0.598} & \tableheat{resultpositive}{17}{0.797} \\
GLM-4.7 & No & 1.2 & \tableheat{resultsmall}{4}{0.272} & \tableheat{resultpositive}{12}{4.554} & \tableheat{resultsmall}{8}{0.493} & \tableheat{resultpositive}{18}{\textbf{2.042}} & \tableheat{resultsmall}{17}{0.939} & \tableheat{resultpositive}{17}{0.883} & \tableheat{resultsmall}{6}{0.205} & \tableheat{resultpositive}{9}{0.318} & \tableheat{resultsmall}{17}{0.592} & \tableheat{resultpositive}{18}{\textbf{0.807}} \\
\addlinespace[1.5pt]
GLM-4.7 & Yes & 0.6 & \tableheat{resultsmall}{15}{0.630} & \tableheat{resultpositive}{17}{6.005} & \tableheat{resultsmall}{7}{0.490} & \tableheat{resultpositive}{12}{1.671} & \tableheat{resultsmall}{18}{0.971} & \tableheat{resultpositive}{16}{0.869} & \tableheat{resultsmall}{18}{\textbf{0.271}} & \tableheat{resultpositive}{4}{0.292} & \tableheat{resultsmall}{18}{\textbf{0.618}} & \tableheat{resultpositive}{16}{0.779} \\
GLM-4.7 & Yes & 1.0 & \tableheat{resultsmall}{16}{0.640} & \tableheat{resultpositive}{17}{6.122} & \tableheat{resultsmall}{8}{0.496} & \tableheat{resultpositive}{15}{1.859} & \tableheat{resultsmall}{18}{\textbf{0.971}} & \tableheat{resultpositive}{17}{0.886} & \tableheat{resultsmall}{16}{0.263} & \tableheat{resultpositive}{4}{0.293} & \tableheat{resultsmall}{18}{0.615} & \tableheat{resultpositive}{17}{0.794} \\
GLM-4.7 & Yes & 1.2 & \tableheat{resultsmall}{16}{0.655} & \tableheat{resultpositive}{18}{\textbf{6.307}} & \tableheat{resultsmall}{9}{0.503} & \tableheat{resultpositive}{17}{1.950} & \tableheat{resultsmall}{18}{0.967} & \tableheat{resultpositive}{18}{\textbf{0.895}} & \tableheat{resultsmall}{14}{0.251} & \tableheat{resultpositive}{5}{0.298} & \tableheat{resultsmall}{18}{0.614} & \tableheat{resultpositive}{18}{0.800} \\
\addlinespace[3pt]
Qw3.6-27B & No & 0.6 & \tableheat{resultsmall}{5}{0.287} & \tableheat{resultpositive}{6}{2.756} & \tableheat{resultsmall}{9}{0.507} & \tableheat{resultpositive}{3}{1.105} & \tableheat{resultsmall}{18}{0.970} & \tableheat{resultpositive}{3}{0.680} & \tableheat{resultsmall}{8}{0.217} & \tableheat{resultpositive}{12}{0.331} & \tableheat{resultsmall}{11}{0.454} & \tableheat{resultpositive}{11}{0.686} \\
Qw3.6-27B & No & 1.0 & \tableheat{resultsmall}{4}{0.265} & \tableheat{resultpositive}{8}{3.353} & \tableheat{resultsmall}{9}{0.508} & \tableheat{resultpositive}{4}{1.155} & \tableheat{resultsmall}{18}{0.966} & \tableheat{resultpositive}{7}{0.740} & \tableheat{resultsmall}{5}{0.196} & \tableheat{resultpositive}{18}{\textbf{0.358}} & \tableheat{resultsmall}{11}{0.448} & \tableheat{resultpositive}{13}{0.721} \\
Qw3.6-27B & No & 1.2 & \tableheat{resultsmall}{4}{0.254} & \tableheat{resultpositive}{9}{3.563} & \tableheat{resultsmall}{10}{0.512} & \tableheat{resultpositive}{4}{1.195} & \tableheat{resultsmall}{18}{0.964} & \tableheat{resultpositive}{9}{0.763} & \tableheat{resultsmall}{3}{0.185} & \tableheat{resultpositive}{18}{0.357} & \tableheat{resultsmall}{10}{0.437} & \tableheat{resultpositive}{13}{0.723} \\
\addlinespace[1.5pt]
Qw3.6-27B & Yes & 0.6 & \tableheat{resultsmall}{16}{0.654} & \tableheat{resultpositive}{15}{5.535} & \tableheat{resultsmall}{9}{0.505} & \tableheat{resultpositive}{5}{1.260} & \tableheat{resultsmall}{12}{0.828} & \tableheat{resultpositive}{11}{0.792} & \tableheat{resultsmall}{17}{0.263} & \tableheat{resultpositive}{11}{0.327} & \tableheat{resultsmall}{9}{0.415} & \tableheat{resultpositive}{9}{0.658} \\
Qw3.6-27B & Yes & 1.0 & \tableheat{resultsmall}{18}{\textbf{0.720}} & \tableheat{resultpositive}{16}{5.806} & \tableheat{resultsmall}{10}{0.516} & \tableheat{resultpositive}{7}{1.367} & \tableheat{resultsmall}{15}{0.900} & \tableheat{resultpositive}{13}{0.816} & \tableheat{resultsmall}{15}{0.253} & \tableheat{resultpositive}{13}{0.334} & \tableheat{resultsmall}{9}{0.400} & \tableheat{resultpositive}{10}{0.667} \\
Qw3.6-27B & Yes & 1.2 & \tableheat{resultsmall}{17}{0.690} & \tableheat{resultpositive}{15}{5.515} & \tableheat{resultsmall}{10}{0.519} & \tableheat{resultpositive}{8}{1.418} & \tableheat{resultsmall}{16}{0.912} & \tableheat{resultpositive}{13}{0.826} & \tableheat{resultsmall}{13}{0.245} & \tableheat{resultpositive}{13}{0.334} & \tableheat{resultsmall}{8}{0.375} & \tableheat{resultpositive}{10}{0.674} \\
\addlinespace[3pt]
Qw3.6-35B & No & 0.6 & \tableheat{resultsmall}{3}{0.235} & \tableheat{resultpositive}{3}{1.950} & \tableheat{resultsmall}{3}{0.454} & \tableheat{resultpositive}{4}{1.165} & \tableheat{resultsmall}{18}{0.967} & \tableheat{resultpositive}{6}{0.730} & \tableheat{resultsmall}{9}{0.222} & \tableheat{resultpositive}{11}{0.323} & \tableheat{resultsmall}{10}{0.444} & \tableheat{resultpositive}{9}{0.645} \\
Qw3.6-35B & No & 1.0 & \tableheat{resultsmall}{3}{0.249} & \tableheat{resultpositive}{5}{2.644} & \tableheat{resultsmall}{3}{0.452} & \tableheat{resultpositive}{5}{1.222} & \tableheat{resultsmall}{18}{0.963} & \tableheat{resultpositive}{9}{0.763} & \tableheat{resultsmall}{6}{0.202} & \tableheat{resultpositive}{14}{0.338} & \tableheat{resultsmall}{9}{0.417} & \tableheat{resultpositive}{10}{0.670} \\
Qw3.6-35B & No & 1.2 & \tableheat{resultsmall}{3}{0.237} & \tableheat{resultpositive}{6}{2.800} & \tableheat{resultsmall}{3}{0.452} & \tableheat{resultpositive}{6}{1.318} & \tableheat{resultsmall}{17}{0.955} & \tableheat{resultpositive}{10}{0.778} & \tableheat{resultsmall}{4}{0.194} & \tableheat{resultpositive}{16}{0.351} & \tableheat{resultsmall}{9}{0.402} & \tableheat{resultpositive}{11}{0.681} \\
\addlinespace[1.5pt]
Qw3.6-35B & Yes & 0.6 & \tableheat{resultsmall}{11}{0.503} & \tableheat{resultpositive}{10}{4.103} & \tableheat{resultsmall}{4}{0.461} & \tableheat{resultpositive}{5}{1.213} & \tableheat{resultsmall}{3}{0.583} & \tableheat{resultpositive}{5}{0.710} & \tableheat{resultsmall}{13}{0.243} & \tableheat{resultpositive}{11}{0.323} & \tableheat{resultsmall}{3}{0.270} & \tableheat{resultpositive}{3}{0.545} \\
Qw3.6-35B & Yes & 1.0 & \tableheat{resultsmall}{15}{0.612} & \tableheat{resultpositive}{12}{4.625} & \tableheat{resultsmall}{5}{0.473} & \tableheat{resultpositive}{6}{1.307} & \tableheat{resultsmall}{9}{0.736} & \tableheat{resultpositive}{9}{0.769} & \tableheat{resultsmall}{11}{0.233} & \tableheat{resultpositive}{12}{0.329} & \tableheat{resultsmall}{12}{0.479} & \tableheat{resultpositive}{11}{0.676} \\
Qw3.6-35B & Yes & 1.2 & \tableheat{resultsmall}{15}{0.626} & \tableheat{resultpositive}{13}{4.736} & \tableheat{resultsmall}{5}{0.472} & \tableheat{resultpositive}{7}{1.371} & \tableheat{resultsmall}{10}{0.760} & \tableheat{resultpositive}{9}{0.771} & \tableheat{resultsmall}{11}{0.229} & \tableheat{resultpositive}{12}{0.332} & \tableheat{resultsmall}{12}{0.476} & \tableheat{resultpositive}{11}{0.684} \\
\bottomrule
\end{tabular}%
}

\end{table}
\clearpage

\subsection{Molecule structural-granularity sensitivity}
\label{app:molgen-granularity}

Bemis--Murcko scaffolds are an intermediate abstraction: canonical molecule identity also preserves
sidechains, whereas a ring-system identity removes both sidechains and the linkers between separate
ring assemblies. We test whether the MolGen result depends on this intermediate choice by applying
all three mappings to the same on-spec molecules. The ring-system key is the sorted collection of
fused or bridged ring assemblies in a molecule; it preserves atom and bond types and repeated
assemblies but ignores how separate assemblies are linked. Acyclic molecules share one empty key.
All columns in Table~\ref{tab:molgen-granularity} use the same 164 prompts, quality gate, and
subset-counting estimator.

At one draw, all three mappings necessarily give the same coverage and select Qwen3.6-27B at
$T=1.0$ with thinking. At $H=20$, exact molecule identity retains that leader, but both
Bemis--Murcko scaffold and ring-system coverage select GLM-4.7 at $T=1.2$ with thinking. The
one-draw leader reaches 3.895 ring systems, versus 4.559 for the ring-system leader, so the
budget-dependent leader change is not specific to the scaffold mapping. The complete ordering is
less stable: scaffold and ring-system coverage have Spearman correlation $0.840$ across the 24
configurations and share two of their Top-5 configurations. For comparison, exact-molecule and
scaffold coverage have correlation $0.909$ and Top-5 overlap four of five. Thus the endpoint winner
and leader reversal survive a coarser structural definition, while finer configuration-rank claims
remain granularity-dependent.

\begin{table}[h]
      \caption{\textbf{MolGen sensitivity to structural granularity.} Entries are the mean expected
            number of distinct on-spec identities over the 164-prompt paper scope. VTC@1 is common to
            all three mappings because one accepted molecule contributes one identity at any granularity.
            Bold marks the largest value in each column.}
      \label{tab:molgen-granularity}
      \centering
      \scriptsize
      \setlength{\tabcolsep}{3pt}
      \resizebox{0.6\linewidth}{!}{
\begin{tabular}{@{}lccrrrr@{}}
\toprule
\multicolumn{3}{c}{\textbf{Configuration}} & \multicolumn{1}{c}{\textbf{One draw}} & \multicolumn{3}{c}{\textbf{Coverage at $H=20$}} \\
\cmidrule(lr){1-3} \cmidrule(lr){4-4} \cmidrule(lr){5-7}
\textbf{Model} & \textbf{Thinking} & $T$ & VTC@1 & Molecule & Scaffold & Ring system \\
\midrule
DS-V4 & No & 0.6 & 0.303 & 4.242 & 3.322 & 2.837 \\
DS-V4 & No & 1.0 & 0.299 & 5.380 & 4.089 & 3.690 \\
DS-V4 & No & 1.2 & 0.298 & 5.664 & 4.362 & 3.996 \\
\addlinespace[1.5pt]
DS-V4 & Yes & 0.6 & 0.618 & 9.338 & 4.344 & 3.123 \\
DS-V4 & Yes & 1.0 & 0.616 & 9.582 & 4.529 & 3.203 \\
DS-V4 & Yes & 1.2 & 0.598 & 9.387 & 4.566 & 3.282 \\
\addlinespace[3pt]
GLM-4.7 & No & 0.6 & 0.326 & 6.016 & 4.648 & 4.268 \\
GLM-4.7 & No & 1.0 & 0.291 & 5.780 & 4.697 & 4.376 \\
GLM-4.7 & No & 1.2 & 0.272 & 5.419 & 4.554 & 4.280 \\
\addlinespace[1.5pt]
GLM-4.7 & Yes & 0.6 & 0.630 & 10.230 & 6.005 & 4.235 \\
GLM-4.7 & Yes & 1.0 & 0.640 & 10.692 & 6.122 & 4.275 \\
GLM-4.7 & Yes & 1.2 & 0.655 & 11.050 & \textbf{6.307} & \textbf{4.559} \\
\addlinespace[3pt]
Qw3.6-27B & No & 0.6 & 0.287 & 3.993 & 2.756 & 2.198 \\
Qw3.6-27B & No & 1.0 & 0.265 & 4.441 & 3.353 & 2.837 \\
Qw3.6-27B & No & 1.2 & 0.254 & 4.520 & 3.563 & 3.061 \\
\addlinespace[1.5pt]
Qw3.6-27B & Yes & 0.6 & 0.654 & 10.608 & 5.535 & 3.731 \\
Qw3.6-27B & Yes & 1.0 & \textbf{0.720} & \textbf{11.397} & 5.806 & 3.895 \\
Qw3.6-27B & Yes & 1.2 & 0.690 & 10.703 & 5.515 & 3.684 \\
\addlinespace[3pt]
Qw3.6-35B & No & 0.6 & 0.235 & 3.395 & 1.950 & 1.811 \\
Qw3.6-35B & No & 1.0 & 0.249 & 4.483 & 2.644 & 2.472 \\
Qw3.6-35B & No & 1.2 & 0.237 & 4.447 & 2.800 & 2.644 \\
\addlinespace[1.5pt]
Qw3.6-35B & Yes & 0.6 & 0.503 & 7.602 & 4.103 & 3.359 \\
Qw3.6-35B & Yes & 1.0 & 0.612 & 9.161 & 4.625 & 3.877 \\
Qw3.6-35B & Yes & 1.2 & 0.626 & 9.458 & 4.736 & 3.859 \\
\bottomrule
\end{tabular}
}
\end{table}
\clearpage
\subsection{Surface-diversity methodology and complete results}
\label{app:surface-diversity}

Surface diversity measures variation that is directly observable in the generated outputs, before
the coarser task-specific mapping used by VTC. For molecule design, we count distinct canonical
molecules that are valid and satisfy the requested property constraints.
For repository repair, we count distinct exact textual diffs among patches that pass
PatchEval-Verified. For bug finding, we count distinct normalized test-input strings that agree
with the reference implementation and kill at least one bug class. For differential diagnosis, we
count distinct canonicalized raw diagnosis terms whose mapped concept set intersects the case's
gold differential. These choices deliberately preserve differences that VTC may merge: for
example, distinct molecules can share a scaffold, distinct patches can modify the same functions,
and distinct diagnosis phrases can map to the same clinical concept.

For these four tasks, we apply the subset-counting estimator in
Equation~\ref{eq:empirical-coverage} to the surface units and average the resulting expected count
equally over task instances. Invalid or off-target attempts contribute no unit but remain in the
attempt budget. Evidence search has no natural exact surface unit. For each saved evidence-selection
attempt, we run a separate answer-generation call using the question and the passages selected in
that attempt. We measure variation among the answers from the first $H=10$ attempts using
$1-\mathrm{SelfBLEU}$. Each answer is compared with the other answers for the same question, and
scores are averaged across answers and questions. We use NLTK 3.10.0 sentence-level BLEU-4 with
method-2 smoothing \citep{chen2014systematic}. Questions with fewer than two nonempty answers receive
surface diversity zero. Higher values indicate greater surface variation, but values are not
comparable across tasks.

Table~\ref{tab:full-surface-diversity-results} pairs each surface measure with VTC at the same
headline budget $H$ for all 24 configurations and uses the same row order as
Table~\ref{tab:full-configuration-results}.

\begin{table}[h]
      \caption{\textbf{Surface diversity and VTC for all 24 configurations.} Each task pairs its
            surface-diversity measure with VTC at the same headline budget $H$. Surface diversity counts
            canonical molecules, accepted exact diffs, exact valid test inputs, and gold-mapped raw diagnosis
            terms; evidence search uses $1-\mathrm{SelfBLEU}$ over retrieval-conditioned answers. Bold marks
            the largest value in each column.}
      \label{tab:full-surface-diversity-results}
      \centering
      \scriptsize
      \setlength{\tabcolsep}{1.5pt}
      \resizebox{\linewidth}{!}{%
\begin{tabular}{@{}lcc*{5}{rr}@{}}
\toprule
\multicolumn{3}{c}{\textbf{Configuration}} & \multicolumn{2}{c}{\textbf{Molecule design}} & \multicolumn{2}{c}{\textbf{Repository repair}} & \multicolumn{2}{c}{\textbf{Bug finding}} & \multicolumn{2}{c}{\textbf{Differential diagnosis}} & \multicolumn{2}{c}{\textbf{Evidence search}} \\
\cmidrule(lr){1-3} \cmidrule(lr){4-5} \cmidrule(lr){6-7}
\cmidrule(lr){8-9} \cmidrule(lr){10-11} \cmidrule(lr){12-13}
\textbf{Model} & \textbf{Thinking} & $T$ & Surf@20 & VTC@20 & Surf@10 & VTC@10 & Surf@5 & VTC@5 & Surf@5 & VTC@5 & Surf@10 & VTC@10 \\
\midrule
DS-V4 & No & 0.6 & 4.24 & 3.32 & 4.82 & 1.45 & 11.79 & 0.76 & 4.29 & 0.31 & 0.36 & 0.75 \\
DS-V4 & No & 1.0 & 5.38 & 4.09 & 5.00 & 1.53 & 12.36 & 0.80 & 4.83 & 0.34 & 0.43 & 0.76 \\
DS-V4 & No & 1.2 & 5.66 & 4.36 & \textbf{5.21} & 1.61 & 12.41 & 0.80 & \textbf{5.02} & 0.34 & 0.47 & 0.77 \\
\addlinespace[1.5pt]
DS-V4 & Yes & 0.6 & 9.34 & 4.34 & 4.83 & 1.45 & 17.01 & 0.84 & 4.10 & 0.34 & 0.38 & 0.75 \\
DS-V4 & Yes & 1.0 & 9.58 & 4.53 & 5.04 & 1.54 & 17.48 & 0.85 & 4.16 & 0.34 & 0.43 & 0.76 \\
DS-V4 & Yes & 1.2 & 9.39 & 4.57 & 5.02 & 1.54 & 17.69 & 0.86 & 4.31 & 0.34 & 0.47 & 0.77 \\
\addlinespace[3pt]
GLM-4.7 & No & 0.6 & 6.02 & 4.65 & 4.53 & 1.88 & 15.06 & 0.83 & 3.10 & 0.29 & 0.37 & 0.78 \\
GLM-4.7 & No & 1.0 & 5.78 & 4.70 & 4.65 & 2.04 & 16.77 & 0.86 & 3.59 & 0.31 & 0.43 & 0.80 \\
GLM-4.7 & No & 1.2 & 5.42 & 4.55 & 4.69 & \textbf{2.04} & 17.17 & 0.88 & 3.81 & 0.32 & 0.47 & \textbf{0.81} \\
\addlinespace[1.5pt]
GLM-4.7 & Yes & 0.6 & 10.23 & 6.01 & 4.37 & 1.67 & 18.72 & 0.87 & 3.03 & 0.29 & 0.36 & 0.78 \\
GLM-4.7 & Yes & 1.0 & 10.69 & 6.12 & 4.54 & 1.86 & 19.65 & 0.89 & 3.11 & 0.29 & 0.40 & 0.79 \\
GLM-4.7 & Yes & 1.2 & 11.05 & \textbf{6.31} & 4.66 & 1.95 & \textbf{20.08} & \textbf{0.89} & 3.25 & 0.30 & 0.43 & 0.80 \\
\addlinespace[3pt]
Qw3.6-27B & No & 0.6 & 3.99 & 2.76 & 3.84 & 1.10 & 11.22 & 0.68 & 3.67 & 0.33 & 0.37 & 0.69 \\
Qw3.6-27B & No & 1.0 & 4.44 & 3.35 & 4.04 & 1.16 & 13.70 & 0.74 & 4.25 & \textbf{0.36} & 0.44 & 0.72 \\
Qw3.6-27B & No & 1.2 & 4.52 & 3.56 & 4.13 & 1.20 & 14.71 & 0.76 & 4.46 & 0.36 & 0.48 & 0.72 \\
\addlinespace[1.5pt]
Qw3.6-27B & Yes & 0.6 & 10.61 & 5.54 & 4.01 & 1.26 & 15.81 & 0.79 & 3.13 & 0.33 & 0.44 & 0.66 \\
Qw3.6-27B & Yes & 1.0 & \textbf{11.40} & 5.81 & 4.25 & 1.37 & 17.34 & 0.82 & 3.30 & 0.33 & 0.50 & 0.67 \\
Qw3.6-27B & Yes & 1.2 & 10.70 & 5.52 & 4.38 & 1.42 & 18.06 & 0.83 & 3.39 & 0.33 & 0.54 & 0.67 \\
\addlinespace[3pt]
Qw3.6-35B & No & 0.6 & 3.39 & 1.95 & 3.66 & 1.17 & 12.31 & 0.73 & 3.71 & 0.32 & 0.38 & 0.65 \\
Qw3.6-35B & No & 1.0 & 4.48 & 2.64 & 3.84 & 1.22 & 13.13 & 0.76 & 4.16 & 0.34 & 0.46 & 0.67 \\
Qw3.6-35B & No & 1.2 & 4.45 & 2.80 & 3.96 & 1.32 & 13.02 & 0.78 & 4.40 & 0.35 & 0.50 & 0.68 \\
\addlinespace[1.5pt]
Qw3.6-35B & Yes & 0.6 & 7.60 & 4.10 & 3.83 & 1.21 & 11.71 & 0.71 & 3.22 & 0.32 & 0.46 & 0.54 \\
Qw3.6-35B & Yes & 1.0 & 9.16 & 4.62 & 4.12 & 1.31 & 14.73 & 0.77 & 3.37 & 0.33 & 0.51 & 0.68 \\
Qw3.6-35B & Yes & 1.2 & 9.46 & 4.74 & 4.22 & 1.37 & 15.17 & 0.77 & 3.53 & 0.33 & \textbf{0.55} & 0.68 \\
\bottomrule
\end{tabular}
      }
\end{table}
\clearpage
\subsection{Instance-resampling robustness of headline comparisons}
\label{app:headline-robustness}

We assess the three headline comparisons that depend most directly on the finite instance set using
10,000 paired bootstrap resamples within each benchmark. Within a task, each
replicate applies the same resampled instance weights to every configuration and metric, then
recomputes the coverage values, rankings, Top-5 sets, regret, and agreement. MolGen is resampled
uniformly over its 164-prompt evaluation set.
We use percentile 95\% intervals throughout; Table~\ref{tab:headline-robustness} summarizes the
resulting robustness checks.

\begin{table}[h]
      \caption{\textbf{Instance-resampling robustness of the headline comparisons.} For leader
            reversals, $\Delta_k$ is VTC@$k$ of the headline-budget leader minus that of the one-attempt
            leader. Proxy-selection results rerank configurations within each resample. Temperature results
            report matched $T=0.6$ versus $T=1.2$ comparisons whose VTC difference remains positive under
            resampling.}
      \label{tab:headline-robustness}
      \centering
\small
\renewcommand{\arraystretch}{1.18}
\setlength{\tabcolsep}{4pt}
\begin{tabularx}{\linewidth}{@{}
>{\raggedright\arraybackslash}p{0.14\linewidth}
>{\raggedright\arraybackslash}p{0.27\linewidth}
>{\raggedright\arraybackslash}p{0.20\linewidth}
>{\raggedright\arraybackslash}X@{}}
\toprule
\textbf{Claim} & \textbf{Scope} & \textbf{Observed result} &
\textbf{Instance-resampling robustness} \\
\midrule
Leader reversal & Molecule design\newline Qw-27B $T{=}1.0$ Tk $\rightarrow$ GLM $T{=}1.2$ Tk & $\Delta_1=-0.066$; $\Delta_H=+0.502$ & $\Delta_1$: [-0.089, -0.044]; $\Delta_H$: [0.150, 0.855] \\
\addlinespace[3pt]
Leader reversal & Repository repair\newline DS $T{=}1.0$ Tk $\rightarrow$ GLM $T{=}1.2$ & $\Delta_1=-0.095$; $\Delta_H=+0.504$ & $\Delta_1$: [-0.129, -0.062]; $\Delta_H$: [0.319, 0.693] \\
\addlinespace[3pt]
Leader reversal & Differential diagnosis\newline DS $T{=}0.6$ Tk $\rightarrow$ Qw-27B $T{=}1.0$ & $\Delta_1=-0.0173$; $\Delta_H=+0.0222$ & $\Delta_1$: [-0.0263, -0.0082]; $\Delta_H$: [0.0092, 0.0352] \\
\addlinespace[3pt]
Leader reversal & Evidence search\newline GLM $T{=}0.6$ $\rightarrow$ GLM $T{=}1.2$ & $\Delta_1=-0.0131$; $\Delta_H=+0.0298$ & $\Delta_1$: [-0.0180, -0.0084]; $\Delta_H$: [0.0222, 0.0376] \\
\addlinespace[3pt]
Proxy selection & One-draw quality\newline five-task macro & regret 12.1\%; agreement 10/25 & regret [10.2\%, 13.9\%]; agreement [8, 12]/25 \\
\addlinespace[3pt]
Proxy selection & Surface diversity\newline five-task macro & regret 11.0\%; agreement 11/25 & regret [9.2\%, 12.4\%]; agreement [9, 13]/25 \\
\addlinespace[3pt]
Temperature & $T=0.6\rightarrow1.2$\newline 40 matched pairs & 38/40 positive & 36/40 intervals above zero\newline MolGen 6/8; PatchEval 7/8; BugFind 8/8; AMIE 7/8; BRIGHT-PRO 8/8 \\
\bottomrule
\end{tabularx}

\end{table}

The intervals for all four fixed point-estimate leader pairs remain below zero at $k=1$ and above
zero at $H$. The task-macro Top-5 regret intervals are 10.2--13.9\% for one-draw quality and
9.2--12.4\% for surface diversity, while their total Top-5 agreement intervals remain 8--12 and
9--13 out of 25, respectively. Of the 38 positive matched temperature comparisons, 36 have
individual intervals entirely above zero. These are
\emph{instance-resampling robustness intervals}: they measure sensitivity to reweighting the frozen
benchmark instances, not uncertainty over newly generated attempts or population-level certainty
over real-world tasks.

\end{document}